\documentclass{article}

\usepackage[preprint]{neurips_2026}

\usepackage[utf8]{inputenc}
\usepackage[T1]{fontenc}
\usepackage{hyperref}
\usepackage{url}
\usepackage{booktabs}
\usepackage{amsfonts}
\usepackage{amsmath}
\usepackage{graphicx}
\usepackage{nicefrac}       
\usepackage{microtype}      
\usepackage{xcolor}
\usepackage{multirow}
\usepackage{array}
\usepackage{caption}
\usepackage{subcaption}
\newenvironment{fixedtable}{%
  \par\medskip
  \noindent\begin{minipage}{\linewidth}
  \captionsetup{type=table}%
}{%
  \end{minipage}
  \par\medskip
}

\newcommand{\metriccell}[2]{\begin{tabular}{@{}c@{}}#1\\{\scriptsize #2}\end{tabular}}

\title{Shallow Prefill, Deep Decoding: Efficient Long-Context Inference via Layer-Asymmetric KV Visibility}
\author{%
  Jungsuk Oh \quad Hyeseo Jeon \quad Hyunjune Ji \quad Kyongmin Kong \quad Jay-Yoon Lee\thanks{Corresponding author} \\\\
  Graduate School of Data Science\\
  Seoul National University\\
  Seoul, Republic of Korea \\
  \texttt{luke0112@snu.ac.kr} \\
}

\begin{document}
\maketitle

\begin{figure}[ht] \centering 
\includegraphics[width=0.95
\linewidth]{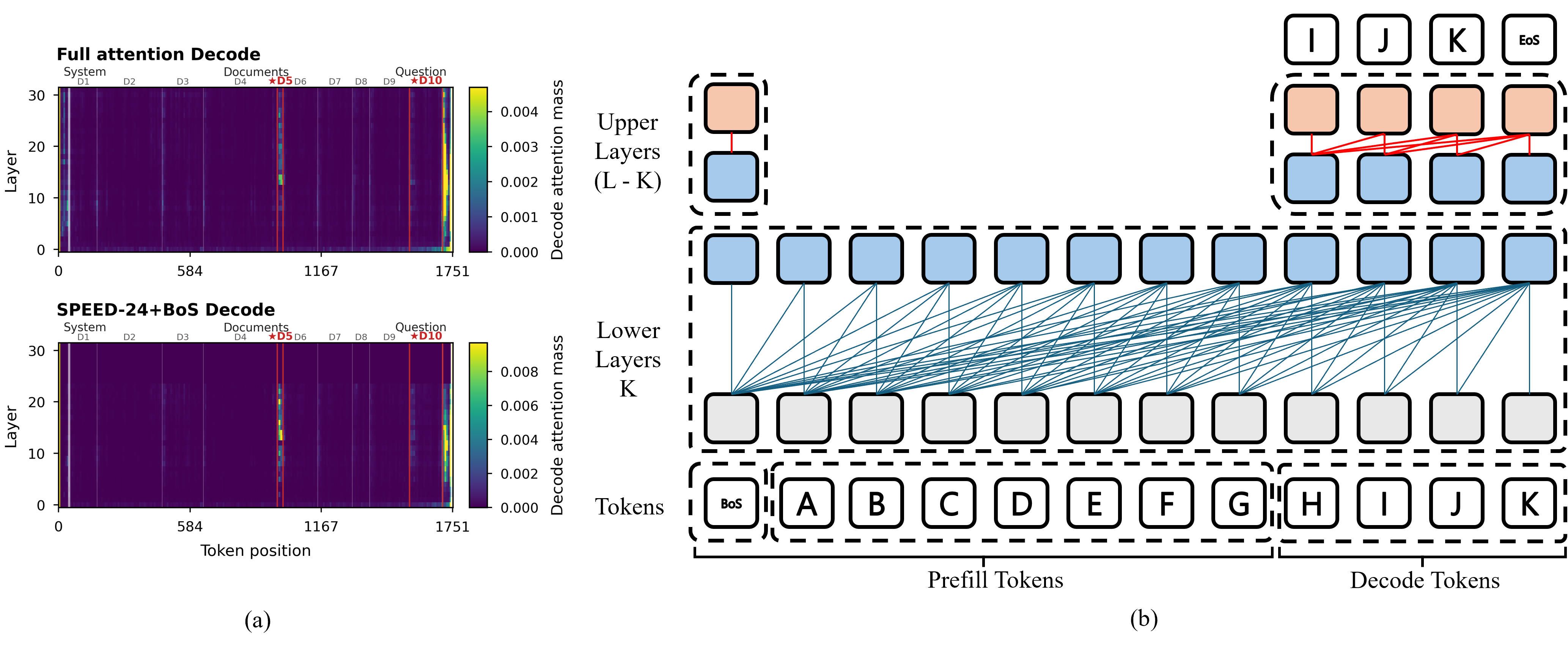} 
\caption{
Overview and attention behavior of SPEED.
Left (a): Decode-time attention-mass heatmaps for Full-Attn (top) and SPEED-24+BoS (bottom), showing that SPEED largely preserves the structured attention pattern of the full-depth model after removing upper-layer prefill-token KV states.
Right (b): SPEED processes prefill tokens only through the first $K$ layers, keeps decode tokens full-depth, and uses the existing BoS token as a stabilization anchor.
}
\label{fig:overview} 
\end{figure}

\begin{abstract}
Long-context inference in decoder-only language models is costly because long prompts are processed during Prefill, cached at every layer, and repeatedly attended to during autoregressive Decode. We introduce \emph{Shallow Prefill, dEEp Decode} (SPEED), a phase-asymmetric KV-visibility policy that materializes non-anchor prompt-token KV states only in lower layers while keeping Decode-phase tokens full-depth. Unlike previous approaches that make upper-layer prompt KV states cheaper to store or construct, SPEED removes prefill tokens from the upper-layer Decode visibility set altogether. With a minimal BoS anchor, this simple change preserves broad benchmark quality while reducing long-context cost. In a controlled Llama-3.1-8B instruction-tuning study, SPEED using only 75\% of layers for prefill tokens reaches 51.2 average score on OLMES-style benchmarks, compared with 51.4 for the full-depth baseline, while improving TTFT by 33\%, TPOT by 22\%, and reducing active KV memory by 25.0\% at 128K context. Layer-wise diagnostics suggest that this cutoff retains the main prompt-selection and representation-stabilization regions of the full-depth model. These results show that long-context prompt tokens need not always persist as full-depth KV-cache objects when Decode-phase tokens remain full-depth.
\end{abstract}

\section{Introduction}

Long-context inference is a central workload for decoder-only language models, including retrieval-augmented generation, document question answering, long-form summarization, and code assistance. In standard autoregressive inference, a model first runs a \emph{Prefill} phase over the input sequence, producing KV states for prefill tokens, and then enters the \emph{Decode} phase, where new tokens are generated one at a time while attending to cached states. In long-context settings, prefill tokens greatly outnumber decode tokens, exposing three coupled costs: Prefill dominates time-to-first-token (TTFT), Decode becomes memory-bandwidth-bound because each new token reads cached KV states, and active KV memory scales with both context length and model depth~\mbox{\citep{pope2023efficiently,patel2024splitwise,zhong2024distserve}}.

Previous research has reduced long-context cost by exploiting redundancy in cached prefill-token states. Some methods make the cache smaller, for example through token selection, cache compression, or quantization~\citep{zhang2023h2o,li2024snapkv,tang2024quest,liu2024kivi}. Others approximate upper-layer KV states by sharing, merging, or transforming representations across depth~\citep{brandon2024reducing,liu2024minicache,qiao2025swiftkv,he2026pop}. These approaches are motivated by a common observation: as layers become deeper, token representations and KV states often become more redundant, and upper-layer attention may contribute less to gathering new prefill-token information than lower-layer attention~\citep{brandon2024reducing,liu2024minicache,artzy2024attend,he2024matters}. The Full-Attn heatmap in Figure~\ref{fig:overview} shows the same intuition in our setting: decode tokens attend strongly to prefill tokens in middle layers, while this prefill-token attention becomes much weaker in upper layers. SPEED pushes this observation further. If lower layers already capture most of the useful prefill-token information, \emph{do we need to keep upper-layer prefill-token KV states in memory for decoding?}

We propose \emph{Shallow Prefill, dEEp Decode} (SPEED), a phase-asymmetric KV-visibility policy that makes prefill tokens shallow while keeping decode tokens deep. In an $L$-layer decoder-only transformer, prefill tokens are processed only through the first $K$ layers, while decode tokens still traverse all $L$ layers and produce full-depth KV states. Thus, lower-layer Decode attention can read the prefill sequence, whereas upper layers attend only to the current decode token and previously generated decode tokens. This reduces long-context cost: for a prefill length $N$, dominant prefill-side KV storage scales as $O(KN)$ rather than $O(LN)$, and upper-layer Decode avoids repeated prefill-cache reads. Following attention-sink observations that initial tokens can stabilize long-context generation~\citep{xiao2023efficient}, we find that the existing BoS token alone is sufficient to stabilize this shallow-Prefill regime. We call this BoS token an anchor, and show that it stabilizes SPEED without restoring upper-layer access to the prefill sequence. Figure~\ref{fig:overview} summarizes the evidence and mechanism: Full-Attn concentrates decode-to-prefill attention in middle layers, SPEED-24+BoS largely preserves this pattern after upper-layer prefill-token KV states are removed, and the overview diagram illustrates the resulting visibility policy.

We evaluate SPEED in two settings. First, we run a controlled instruction-tuning sweep from Llama-3.1-8B Base~\citep{grattafiori2024llama}, where the full-depth instruction-tuned baseline (\emph{Full-IT}) and all SPEED variants share the same data, formatting, optimizer, and evaluation protocol, isolating the effect of KV visibility. Our main operating point, SPEED with $K=24$ and BoS anchoring (\emph{SPEED-24+BoS}), uses only 75\% of layers for prefill tokens and reaches 51.2 average score across OLMES-style benchmarks~\citep{gu2025olmes}, compared with 51.4 for Full-IT. At 128K context, it improves TTFT by 33\%, TPOT by 22\%, and reduces active KV memory by 25.0\%. BoS anchoring is also important: at $K=24$, it raises the average score from 49.1 to 51.2 without changing the efficiency profile. Second, to test a lower-cost adaptation path, we start from an off-the-shelf Llama-3.1-8B-Instruct checkpoint and apply one epoch of low-rank adaptation (\emph{LoRA}). Moderate SPEED cutoffs remain competitive with full-depth LoRA adaptation on document-grounded QA and long-context retrieval, showing that SPEED can also be applied through lightweight adaptation. We further provide layer-wise diagnostics that connect the quality--efficiency frontier to prefill-token selectivity and representation stabilization in the full-depth model.

\paragraph{Contributions.}
\begin{itemize}
    \item We introduce \emph{SPEED}, a phase-asymmetric KV-visibility policy that makes prefill tokens shallow while keeping Decode-phase tokens full-depth, thereby removing upper-layer prefill-token KV states without reducing Decode depth.

    \item We show that SPEED-24+BoS is a strong operating point: using only 75\% of layers for prefill tokens, it remains close to the full-depth instruction-tuned baseline while reducing 128K-context TTFT, TPOT, and active KV memory.

    \item We demonstrate that a single BoS anchor is sufficient to stabilize the shallow-Prefill regime, and that SPEED can also be applied through one epoch of LoRA adaptation from an off-the-shelf instruction model.

    \item We provide a layer-wise cutoff diagnostic that helps guide the choice of $K$, reducing reliance on exhaustive cutoff sweeps by tracking prefill-token selectivity, attention to previously generated decode tokens, and representation stabilization in the full-depth model.
\end{itemize}
\begin{figure}[t]
    \centering
    \includegraphics[width=\linewidth]{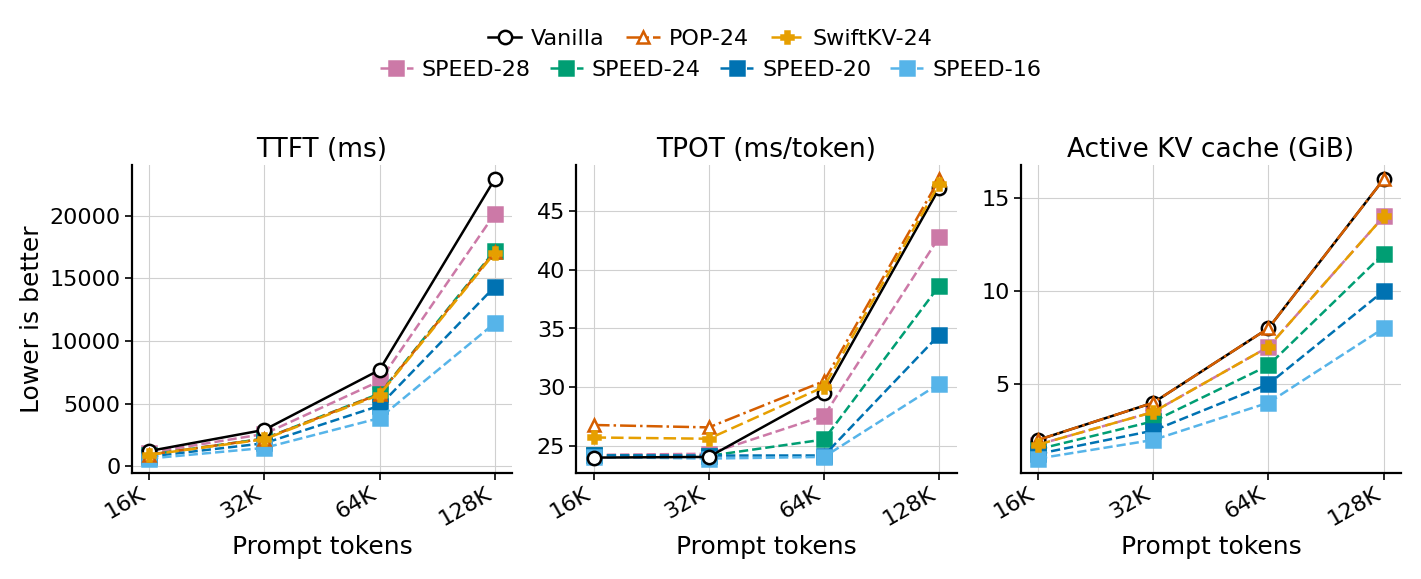}
    \caption{
    Long-context efficiency on Llama-3.1-8B with a fixed 128-token continuation.
    We compare Full-Attn, SPEED cutoffs, SwiftKV-24, and POP-24 under the same measurement protocol.
    }
    \label{fig:efficiency}
\end{figure}

\section{Related Work}

\paragraph{KV-cache reduction and serving systems.}
Long-context inference has been accelerated by reducing how many KV states are stored, how many bytes each state occupies, or how much KV traffic is incurred during attention and serving. Token-selection and eviction methods retain recent, heavy-hitter, or query-relevant KV states~\citep{zhang2023h2o, li2024snapkv, tang2024quest}, while KV quantization reduces the memory footprint of each cached key and value~\citep{liu2024kivi}. Sparse-attention, head-wise routing, and serving systems further reduce attention computation, KV traffic, or cache-management overhead through structured sparsity, selective full-context access, paging, and virtualized allocation~\citep{jiang2024minference, xiao2024duoattention, kwon2023efficient, prabhu2025vattention}. Recent KV-admission work asks which token states should be written into persistent memory in the first place~\citep{huang2025kv}. SPEED is related to this admission perspective, but differs in mechanism: it does not perform online token scoring, eviction, compression, routing, or learned admission. Once the cutoff $K$ and anchor set are fixed, non-anchor prefill tokens are processed in lower layers but are never materialized as upper-layer KV objects.

\paragraph{Depth-wise KV reduction and phase-aware Prefill optimization.}
SPEED is most closely related to methods that exploit redundancy across transformer depth or asymmetry between Prefill and Decode. Depth-wise KV methods share, merge, condense, or allocate KV budgets across layers~\citep{wu2024layer, brandon2024reducing, sun2024you, liu2024minicache, cai2024pyramidkv, dehghanighobadi2026depthkv}. Stage-aware Prefill methods are especially close. SwiftKV constructs later-layer KV caches from earlier representations and merges neighboring-layer caches~\citep{qiao2025swiftkv}, while POP removes deep-layer computation during Prefill while retaining full-depth Decode through independent KV projections and boundary handling~\citep{he2026pop}. These approaches reduce or restructure Prefill-side work, but still preserve, share, or synthesize upper-layer prefill-token KV states for Decode. Figure~\ref{fig:efficiency} highlights the consequence under our measurement protocol: at the comparable $K=24$ operating point, POP-24, SwiftKV-24, and SPEED-24 obtain similar TTFT reductions, but only SPEED-24 improves TPOT and yields the lowest active KV memory. SPEED therefore differs not by merely accelerating Prefill, but by changing the Decode-time visibility set itself: non-anchor prefill tokens are absent from upper-layer Decode attention, reducing repeated upper-layer prefill-cache reads during autoregressive generation.

\paragraph{Depth-adaptive inference, prompt surrogates, and layer-wise roles.}
Early-exit, layer-skipping, and pruning methods reduce computation by allowing examples, tokens, heads, or layers to bypass part of the model~\citep{fan2019reducing, schuster2022confident, elhoushi2024layerskip, he2024matters, liu2025high, saikumar2025data}. SPEED is different: Decode tokens still traverse all layers and produce full-depth KV states, so it is not early exiting generation. Prompt-compression and learned-surrogate methods construct compact input representations, such as gist tokens or compressed context embeddings~\citep{mu2023learning, chevalier2023adapting, ge2023context}. SPEED instead retains direct prompt access in lower layers while removing non-anchor prefill-token KV materialization from upper layers. SPEED+BoS is motivated by attention-sink observations that initial tokens can stabilize long-context generation~\citep{xiao2023efficient}. More broadly, analyses of layer-wise behavior suggest that attention, information selection, and representation formation vary across depth~\citep{artzy2024attend, hosseini2023large}. SPEED turns this layer-wise asymmetry into a prefill-depth allocation policy: preserve full-depth Decode computation, but reduce the depth at which prefill tokens persist as cached memory.

\section{SPEED: Shallow Prefill, dEEp Decode}
\label{sec:method}

SPEED is a layer-wise KV-visibility policy for decoder-only transformers~\citep{vaswani2017attention}. It keeps Decode-phase tokens full-depth while making non-anchor prefill-token KV materialization shallow. In an $L$-layer model with cutoff $K<L$, non-anchor prefill tokens are processed and cached only through layers $\{1,\ldots,K\}$, whereas Decode-phase tokens traverse all $L$ layers and produce full-depth KV states for future generation. Optional anchors, such as BoS, are retained through all layers. Thus, SPEED changes KV visibility, not the transformer weights, language-modeling objective, or positional indices.

\paragraph{Token visibility.}
Let $s$ denote the BoS token, $X$ the remaining non-BoS prefill tokens, $D_{<t}$ previous Decode-phase tokens, and $d_t$ the current Decode-phase token. We define a prefill-side anchor as a prefill token whose KV states are materialized through all $L$ layers and remain visible to upper-layer Decode attention. Anchor-free SPEED uses no prefill-side anchor, while SPEED+BoS uses the existing BoS token as the only full-depth prefill-side anchor. BoS is not a learned summary, compressed prompt representation, or additional memory token; it is a minimal stable reference retained from the original sequence.

For the current Decode-phase token $d_t$, Table~\ref{tab:visibility-policies} summarizes the visible KV set at lower and upper layers. The key distinction is that Decode-phase tokens remain full-depth in all SPEED variants. Only non-anchor prefill-token KV materialization is truncated.

\begin{table}[t]
\caption{
Layer-wise visible KV sets for the current Decode-phase token $d_t$.
Anchor-free SPEED removes all prefill-side upper-layer KV states; SPEED+BoS retains only BoS as a full-depth prefill-side anchor.
}
\label{tab:visibility-policies}
\centering
\setlength{\tabcolsep}{3.5pt}
\begin{tabular}{lll}
\toprule
Policy & Lower layers $(l \le K)$ & Upper layers $(l>K)$ \\
\midrule
Full-Attn &
$X \cup \{s\} \cup D_{<t} \cup \{d_t\}$ &
$X \cup \{s\} \cup D_{<t} \cup \{d_t\}$ \\
Anchor-free SPEED &
$X \cup \{s\} \cup D_{<t} \cup \{d_t\}$ &
$D_{<t} \cup \{d_t\}$ \\
SPEED+BoS &
$X \cup \{s\} \cup D_{<t} \cup \{d_t\}$ &
$\{s\} \cup D_{<t} \cup \{d_t\}$ \\
\bottomrule
\end{tabular}
\end{table}

Anchor-free SPEED cleanly exposes the no-upper-prefill-KV regime, but it can destabilize generation when early Decode steps have very small upper-layer key sets. SPEED+BoS is therefore our main stabilized variant: it adds only one full-depth prefill-side KV state while leaving all other prefill tokens lower-layer-only.

\paragraph{Cost model.}
Let $P$ be the set of non-anchor prefill tokens and let $N=|P|$. In SPEED+BoS, the anchor set is $A=\{s\}$ and $P=X$; in anchor-free SPEED, $A=\emptyset$ and $P=X\cup\{s\}$. Let $a=|A|$ be the number of full-depth prefill-side anchors, and let $T=|D_{<t}|$ be the number of cached Decode-phase tokens. Finally, let
\begin{equation}
B_{\mathrm{KV}} = 2 n_{\mathrm{kv}} d_{\mathrm{head}} b
\end{equation}
be the bytes required for one token's key and value at one layer. Full attention stores every prefill and Decode-phase token at every layer:
\begin{equation}
M_{\mathrm{Full}} \approx B_{\mathrm{KV}} L(N+a+T).
\end{equation}
SPEED stores non-anchor prefill tokens only in the first $K$ layers, while anchors and Decode-phase tokens remain full-depth:
\begin{equation}
M_{\mathrm{SPEED}} \approx B_{\mathrm{KV}}(KN+La+LT).
\end{equation}
Thus, for long prompts where $N\gg a,T$, the dominant prefill-side KV memory is reduced from $O(LN)$ to $O(KN)$. The same layer-token reduction applies to Prefill computation and to the prefill-token portion of Decode-time attention:
\begin{equation}
C_{\mathrm{prefill}}, R_{\mathrm{decode}}:
\quad
L(N+a) \;\rightarrow\; KN+La.
\end{equation}
These expressions are scaling proxies rather than a complete latency model; realized TTFT and TPOT also depend on kernels, memory bandwidth, cache layout, batching, and serving implementation.

\paragraph{Training and implementation.}
During SPEED-aware supervised fine-tuning, prompt positions follow the prefill-token visibility rule, while assistant target positions follow the Decode-token rule under teacher forcing. The loss and target tokens are unchanged. We implement SPEED by controlling KV-cache materialization and layer-wise attention visibility: non-anchor prefill-token KV tensors are materialized only for layers $1$ through $K$, while anchor tokens and Decode-phase tokens are materialized at all layers. Position indices are not renumbered, so SPEED changes which KV states are visible, not token positional identity.

\section{Experimental Setup}
\label{sec:setup}

All main experiments use the 32-layer Llama-3.1-8B architecture~\citep{grattafiori2024llama}. We evaluate prefill-visible cutoffs $K\in\{16,20,24,28\}$, with $K=32$ corresponding to standard full-depth attention. Our primary comparison is a controlled instruction-tuning study from Llama-3.1-8B Base. The full-depth baseline and all SPEED variants use the same supervised fine-tuning mixture, chat formatting, optimizer, learning-rate schedule, batch construction, and number of updates; the intended difference is the layer-wise KV-visibility policy. The instruction-tuning mixture contains 178{,}502 examples, corresponding to a 20\% subsample of a Tulu-style supervised fine-tuning mixture~\citep{lambert2024tulu}, and each model is trained for two epochs.

We denote the full-depth instruction-tuned model as \emph{Full-IT}, anchor-free SPEED models as \emph{IT-SPEED-$K$}, and BoS-anchored models as \emph{IT-SPEED-$K$+BoS}. IT-SPEED-$K$+BoS is our main method; anchor-free IT-SPEED-$K$ is used as a diagnostic setting. Detailed hyperparameters, task identifiers, hardware, and inference configurations are provided in Appendix~\ref{app:setup-details}.

\paragraph{General-capability and efficiency evaluation.}
We evaluate instruction-tuned quality on TULU-3-DEV under the OLMES-style protocol~\citep{gu2025olmes}. We report the unweighted macro-average over 11 benchmark scores and five category aggregates: Knowledge, Reasoning, Code, Math, and Instruction. Category definitions and full per-benchmark results are provided in Appendix~\ref{app:full-results}.

For long-context efficiency, we measure prompt lengths from 1K to 128K tokens with a fixed 128-token continuation, repeating each setting five times. We report TTFT, TPOT, active KV-cache memory, and estimated FLOPs. Speedups and memory reductions are computed relative to the full-depth $K=32$ baseline under the same inference configuration. Active KV memory counts materialized KV tensors, and FLOPs are estimated from the layer-token scaling proxy in Section~\ref{sec:method}. We include POP-24 and SwiftKV-24 as efficiency-only stage-aware Prefill baselines; these baselines are used for efficiency comparison only, not for matched general-capability quality comparison.

\paragraph{Off-the-shelf LoRA compatibility.}
Because full instruction tuning from a base checkpoint can be costly, we also test a lighter adaptation path. Starting from Llama-3.1-8B-Instruct, we apply one epoch of LoRA task adaptation on HotpotQA pseudo-labeled training examples and evaluate document-grounded QA transfer and synthetic long-context retrieval. We compare SPEED+BoS LoRA adaptation with full-depth LoRA adaptation under the same task-adaptation setup. Additional task-adaptive results are provided in Appendix~\ref{app:taskft-full}.

\paragraph{Layer-wise cutoff diagnostic.}
To guide cutoff selection, we run layer-wise diagnostics on Full-IT using TULU-3-DEV prompts. During greedy Decode, we measure attention from generated Decode-phase tokens to prefill tokens, BoS, and earlier Decode-phase tokens. We also compute conditional prompt entropy over prefill tokens and hidden-trajectory straightening as a representation-stabilization signal~\citep{henaff2021primary,hosseini2023large}. These diagnostics are used to interpret where prefill visibility can be reduced, not as causal proofs of layer roles or per-example cutoff predictors. Sampling and filtering details are provided in Appendix~\ref{app:layer-diagnostic-setup}.

\paragraph{Upper-layer Decode-token attention ablation.}
To test whether SPEED can also remove upper-layer attention among Decode-phase tokens, we evaluate a SelfOnly diagnostic variant. SelfOnly follows the same shallow-Prefill visibility rule as SPEED, but upper-layer Decode-phase tokens attend only to their own current position, optionally with a BoS anchor, rather than attending to other Decode-phase tokens. This ablation tests what SPEED preserves: full-depth Decode computation and upper-layer Decode-token attention. Full SelfOnly results are provided in Appendix~\ref{app:selfonly}.

\paragraph{Additional checks.}
Appendix experiments cover task-adaptive transfer, length robustness, repetition-loop analysis, additional SelfOnly variants, and training throughput. We use these as supporting evidence and failure-mode analysis rather than as the primary basis for the quality--efficiency frontier.

\section{Results}
\label{sec:results}

We present four sets of results. First, we show that BoS anchoring recovers most of the quality loss from shallow Prefill while preserving SPEED's TTFT, TPOT, and KV-memory gains. Second, we show that SPEED can be introduced through lightweight LoRA adaptation from an off-the-shelf instruction model. Third, we analyze why $K=24$ is a useful cutoff and why more aggressive cutoffs degrade. Finally, we summarize additional appendix experiments that test robustness and clarify failure modes.

\subsection{BoS anchoring yields a strong quality--efficiency point}
\label{sec:main-frontier}

Table~\ref{tab:main-general-efficiency} reports category-level general capability and 128K-context efficiency after controlled instruction tuning. Efficiency numbers are computed relative to Full-IT under the same inference configuration. Figure~\ref{fig:efficiency} compares long-context efficiency against the efficiency-only POP-24 and SwiftKV-24 baselines across prompt lengths.

\begin{table}[t]
\caption{
General capability and 128K-context efficiency after SPEED-aware instruction tuning.
TTFT and TPOT report speedup percentages relative to Full-IT.
KV reports active KV-cache memory reduction relative to Full-IT.
}
\label{tab:main-general-efficiency}
\centering
\setlength{\tabcolsep}{3.0pt}
\begin{tabular}{lcccccccc}
\toprule
Method & Avg. & Know. & Reason. & Code & Math & Inst. & TTFT & TPOT / KV \\
\midrule
Full-IT ($K=32$) & \textbf{51.4} & 44.9 & 57.8 & 74.6 & \textbf{46.6} & \textbf{36.1} & -- & -- / -- \\
\midrule
IT-SPEED-28 & 50.2 & 44.0 & 56.1 & 73.8 & 45.2 & 35.0 & +14\% & +10\% / 12.5\% \\
IT-SPEED-28+BoS & 51.3 & 45.7 & 57.0 & 75.2 & \textbf{46.6} & 34.9 & +14\% & +10\% / 12.5\% \\
\midrule
IT-SPEED-24 & 49.1 & 41.7 & 54.8 & \textbf{75.6} & 43.1 & 34.3 & +33\% & +22\% / 25.0\% \\
IT-SPEED-24+BoS & 51.2 & \textbf{46.0} & \textbf{58.0} & 75.4 & 45.3 & 33.9 & +33\% & +22\% / 25.0\% \\
\midrule
IT-SPEED-20 & 48.6 & 42.3 & 55.6 & 73.6 & 42.8 & 32.0 & +60\% & +36\% / 37.5\% \\
IT-SPEED-20+BoS & 49.9 & 45.1 & 56.1 & 75.5 & 42.9 & 32.3 & +60\% & +36\% / 37.5\% \\
\midrule
IT-SPEED-16 & 44.3 & 39.3 & 45.6 & 74.1 & 36.1 & 29.1 & +101\% & +55\% / 50.0\% \\
IT-SPEED-16+BoS & 45.4 & 43.0 & 44.0 & 71.0 & 40.6 & 29.9 & +101\% & +55\% / 50.0\% \\
\bottomrule
\end{tabular}
\end{table}

At $K=24$, anchor-free SPEED drops from 51.4 to 49.1 average score, showing that removing upper-layer prefill-token KV states without a stable prefill-side reference can hurt quality. Adding the BoS anchor recovers most of this loss: IT-SPEED-24+BoS reaches 51.2 average score, only 0.2 points below Full-IT. The same stabilization appears in the repetition analysis in Appendix~\ref{app:repetition-analysis}, where BoS anchoring suppresses suffix-repetition loops observed in anchor-free SPEED. The efficiency profile is unchanged by the anchor. At 128K context, IT-SPEED-24+BoS improves TTFT by 33\%, improves TPOT by 22\%, and reduces active KV memory by 25.0\%. Thus, in this setting, a single full-depth BoS token stabilizes shallow Prefill without restoring upper-layer access to the full prefill sequence.

The cutoff sweep also shows that quality degradation is task-dependent. Code is relatively robust to shallow prefill visibility: its score remains close to Full-IT under moderate cutoffs, and even the anchor-free $K=16$ setting stays near the full-depth code score. Math and Instruction are more sensitive. Math drops sharply at $K=16$ and recovers only at moderate cutoffs, while Instruction declines steadily as $K$ decreases. Knowledge and Reasoning benefit substantially from BoS anchoring at moderate cutoffs, suggesting that these categories need a stable upper-layer prefill-side reference but not necessarily full-depth KV states for all prefill tokens. This pattern motivates the layer-wise diagnostic in Section~\ref{sec:layer-diagnostics-main}, where we examine where prompt selection and representation stabilization occur across depth.

Figure~\ref{fig:efficiency} complements the quality results by comparing SPEED with stage-aware Prefill baselines. At the comparable $K=24$ operating point, SPEED-24, POP-24, and SwiftKV-24 obtain similar TTFT reductions, indicating that all three reduce or restructure Prefill-side work. The difference appears during Decode. POP-24 and SwiftKV-24 do not improve TPOT over Full-Attn in our implementation, and their active KV footprints remain larger than SPEED-24. SPEED improves TPOT because non-anchor prefill tokens are absent from upper-layer Decode attention, reducing repeated upper-layer prefill-cache reads during autoregressive generation.

\subsection{SPEED can be adapted from an off-the-shelf instruction model}
\label{sec:offshelf-main}

\begin{table}[t]
\caption{
Off-the-shelf instruction-model compatibility with lightweight SPEED adaptation.
All adapted rows start from Llama-3.1-8B-Instruct and use one epoch of LoRA task adaptation on HotpotQA pseudo-labeled training examples.
Full-depth LoRA denotes full-depth adaptation under the same setup.
QA columns report EM/F1; S-NIAH reports exact match.
}
\label{tab:offshelf-main}
\centering
\small
\setlength{\tabcolsep}{3.8pt}
\begin{tabular}{lcccc}
\toprule
Method & HotpotQA & TriviaQA & NQ & S-NIAH \\
\midrule
Llama3.1 8B Instruct & 56.9 / 72.7 & 78.8 / 84.8 & 45.8 / 61.1 & 99.6 \\
Full-depth LoRA & 60.8 / 75.3 & 80.5 / 86.0 & 48.5 / 62.4 & 97.7 \\
OffShelf-FT-SPEED+BoS-28 & 58.7 / 73.4 & 81.3 / 86.5 & 47.9 / 61.5 & 97.0 \\
OffShelf-FT-SPEED+BoS-24 & 59.5 / 73.7 & 81.4 / 86.5 & 46.4 / 59.8 & 99.6 \\
OffShelf-FT-SPEED+BoS-20 & 59.4 / 73.5 & 81.1 / 86.4 & 45.4 / 58.7 & 96.1 \\
OffShelf-FT-SPEED+BoS-16 & 55.0 / 69.4 & 76.7 / 81.7 & 39.2 / 52.8 & 88.8 \\
\bottomrule
\end{tabular}
\end{table}

The controlled Base-to-SFT sweep isolates SPEED under matched instruction-tuning conditions, but it requires training from a base checkpoint. Table~\ref{tab:offshelf-main} tests a cheaper path: applying SPEED through one epoch of LoRA adaptation from an already instruction-tuned Llama-3.1-8B-Instruct model. Moderate SPEED cutoffs remain close to full-depth LoRA. In particular, OffShelf-FT-SPEED+BoS-24 reaches 59.5/73.7 on HotpotQA, 81.4/86.5 on TriviaQA, and 99.6 on S-NIAH, compared with 60.8/75.3, 80.5/86.0, and 97.7 for full-depth LoRA. Since the adaptation data come from HotpotQA, the TriviaQA and S-NIAH results suggest that SPEED adaptation preserves document-grounded transfer and long-context retrieval behavior rather than only fitting the adaptation task. Additional task-adaptive and off-the-shelf results are reported in the appendix.

\subsection{Layer-wise diagnostics guide cutoff selection}
\label{sec:layer-diagnostics-main}

The main frontier raises a cutoff-selection question: why does $K=24$ preserve quality while more aggressive cutoffs degrade? We analyze the full-depth model to locate where prompt access and representation stabilization occur across layers. For each category, we measure attention from generated Decode-phase tokens to prefill tokens, BoS, and earlier Decode-phase tokens. We also compute conditional prompt entropy over prefill tokens and hidden-trajectory straightening as a representation-stabilization signal. Prompt-attention mass indicates where generated tokens attend to the prompt; conditional prompt entropy indicates how selective that prompt access is; and straightening measures where hidden trajectories become more geometrically stable across layers. Table~\ref{tab:layer-diagnostics-main} summarizes the category-level peaks.

\begin{table}[t]
\caption{
Layer-wise diagnostic peaks on Full-IT using TULU-3-DEV prompts.
Layer indices are 1-based.
Entropy min denotes the minimum conditional prompt entropy over prefill tokens; Straight. peak denotes maximum hidden-trajectory straightening.
}
\label{tab:layer-diagnostics-main}
\centering
\small
\setlength{\tabcolsep}{4.0pt}
\begin{tabular}{lccccc}
\toprule
Category & $n$ & Prompt peak & Decode-token peak & Entropy min & Straight. peak \\
\midrule
Math & 200 & L14 & L13 & L15 & L19 \\
Coding & 200 & L3 & L13 & L3 & L19 \\
Reasoning & 200 & L1 & L13 & L14 & L18 \\
Knowledge & 300 & L1 & L13 & L13 & L17 \\
Instruction & 200 & L14 & L1 & L15 & L19 \\
\bottomrule
\end{tabular}
\end{table}

The diagnostic shows that raw prompt-attention mass alone is not a reliable cutoff signal. Reasoning and Knowledge have prompt-mass peaks at L1, but their conditional-entropy minima occur much later, around L13--L14. This suggests that early layers may attend broadly to prefill tokens, while selective prompt access emerges in middle layers. The straightening peaks occur later still, typically around L17--L19, indicating a subsequent representation-stabilization region.

This pattern explains why very shallow cutoffs can fail even when they include some high-attention layers. A cutoff near L16 may capture parts of prompt access, but leaves little prompt-visible computation after selective prompt use and before stabilization. In contrast, $K=24$ covers the middle-layer selection region and leaves several prompt-visible layers beyond the observed straightening peaks. This gives a practical rule for broad settings: choose $K$ above the layers where selective prompt access and representation stabilization occur, rather than from raw prompt-attention mass alone.

Coding is the main exception. Its prompt-mass peak and entropy minimum both occur at L3, while straightening still peaks at L19. This matches the cutoff sweep, where code scores remain relatively robust even under aggressive truncation. We therefore interpret the diagnostic as evidence that required prefill-visible depth is task-dependent, not that a single cutoff is universally optimal. Full diagnostic summaries, including BoS-related measurements and correlation statistics, are provided in Appendix~\ref{app:layer-diagnostics}.

\subsection{Upper-layer Decode-token attention remains necessary}
\label{sec:selfonly-main}

SPEED removes upper-layer KV states for prefill tokens, but it does not remove upper-layer attention among Decode-phase tokens. We test whether this Decode-token attention is necessary with a SelfOnly diagnostic variant. SelfOnly follows the same shallow-Prefill visibility rule as SPEED, but upper-layer Decode-phase tokens attend only to their own current position, optionally with a BoS anchor, rather than attending to other Decode-phase tokens.

\begin{table}[t]
\caption{
Upper-layer Decode-token attention ablation.
SelfOnly removes upper-layer attention to other Decode-phase tokens while keeping the same shallow-Prefill visibility rule.
}
\label{tab:selfonly-main}
\centering
\small
\setlength{\tabcolsep}{3.5pt}
\begin{tabular}{lcccccc}
\toprule
Method & Avg. & Know. & Reason. & Code & Math & Inst. \\
\midrule
Full-IT & \textbf{51.4} & 44.9 & 57.8 & 74.6 & \textbf{46.6} & \textbf{36.1} \\
IT-SPEED-24+BoS & 51.2 & \textbf{46.0} & \textbf{58.0} & \textbf{75.4} & 45.3 & 33.9 \\
SelfOnly-24+BoS & 47.2 & 40.6 & 55.3 & 71.0 & 41.5 & 31.3 \\
\bottomrule
\end{tabular}
\end{table}

Table~\ref{tab:selfonly-main} shows that upper-layer Decode-token attention is not redundant. SelfOnly-24+BoS drops to 47.2 average score, compared with 51.2 for IT-SPEED-24+BoS, with degradation across all reported categories. This ablation clarifies what SPEED removes and what it preserves. SPEED removes upper-layer access to prefill-token KV states, but Decode-phase tokens still need upper-layer interaction with other Decode-phase tokens. In other words, the efficiency gain should not be interpreted as evidence that upper-layer attention is unnecessary. It comes from removing the long prefill sequence from the upper-layer Decode visibility set while preserving full-depth Decode computation and Decode-token attention. Full SelfOnly results, including conservative cutoffs and anchor-free variants, are reported in Appendix~\ref{app:selfonly}.

\subsection{Additional experiments are reported in the appendix}
\label{sec:secondary-robustness}

The appendix reports supporting experiments that test robustness beyond the controlled instruction-tuning frontier and clarify failure modes of more aggressive visibility restrictions. Appendix~\ref{app:taskft-full} reports task-adaptive transfer after downstream fine-tuning. Across document QA, summarization, math, and code, moderate SPEED+BoS cutoffs remain close to the corresponding full-depth task-adapted baselines while retaining active-KV-memory savings. We treat these results as secondary because they use task-specific adaptation data and cover fewer settings than the controlled instruction-tuning sweep.

Appendix~\ref{app:length-robustness} evaluates long-context length robustness on TriviaQA and S-NIAH. Appendix~\ref{app:repetition-analysis} analyzes suffix-repetition loops and shows that BoS anchoring suppresses an instability observed in anchor-free SPEED. Appendix~\ref{app:selfonly} provides the full SelfOnly diagnostic results, including additional cutoffs and anchor-free variants, complementing the main Decode-token attention ablation. Appendix~\ref{app:training-efficiency} reports training-throughput measurements. Together, these experiments support the same interpretation as the main results: moderate SPEED+BoS cutoffs preserve practical quality while reducing long-context inference cost, whereas more aggressive restrictions expose task-dependent failure modes.

\section{Limitations}
\label{sec:limitations}

SPEED changes the information available to upper layers, rather than merely changing cache layout or memory allocation. Its behavior therefore depends on the input-visible cutoff $K$, anchor design, adaptation procedure, prompt and continuation lengths, task distribution, and model architecture. Aggressive cutoffs can degrade quality, and anchor-free SPEED can destabilize generation by removing all full-depth prefill-side anchors. We therefore use SPEED+BoS as the main stabilized variant and treat anchor-free SPEED as a diagnostic setting. This work evaluates fixed cutoffs $K\in\{16,20,24,28\}$, a minimal BoS anchor, and the 32-layer Llama-3.1-8B architecture; adaptive visibility policies, alternative anchors, other model families, and other scales may produce different cutoff frontiers.

Our broad quality results are controlled matched-run evaluations, not statistical equivalence tests. Small aggregate gaps should therefore be interpreted as evidence from this experimental setting, not as proof that shallow Prefill is lossless. The task-adaptive and off-the-shelf LoRA experiments show compatibility beyond the main controlled instruction-tuning sweep, but they do not exhaust all long-context task distributions or deployment settings. Similarly, the layer-wise diagnostics provide guidance for choosing $K$ by identifying where prompt selection and representation stabilization occur, but they are not causal proofs of layer roles or reliable per-example cutoff predictors.

Measured efficiency gains also depend on the serving stack. Our cost model captures the dominant scaling terms for Prefill computation, Decode-time prefill-token attention, and active KV memory, but realized TTFT and TPOT also depend on kernels, cache layout, batching, CUDA graphs, memory bandwidth, and KV-cache managers. SPEED should therefore be evaluated under the target deployment configuration, especially with continuous batching, prefix sharing, speculative decoding, or custom serving systems. Our POP and SwiftKV comparisons are efficiency-only comparisons under a shared measurement protocol; we do not claim quality dominance over those systems without matched quality evaluations.

\section{Conclusion}

We introduced SPEED, a phase-asymmetric KV-visibility policy that makes Prefill shallow while keeping Decode deep. Prefill tokens are processed and cached only through a lower-layer prefix, while Decode-phase tokens still traverse all layers and produce full-depth KV states. A minimal BoS anchor stabilizes this regime without restoring upper-layer access to the full prefill sequence. On Llama-3.1-8B, SPEED+BoS forms a practical quality--efficiency point: the $K=24$ setting remains close to Full-IT quality while achieving a 33\% TTFT speedup, a 22\% TPOT speedup, and a 25.0\% active-KV-memory reduction at 128K context. These results suggest that long-context efficiency can be improved not only by compressing, selecting, or serving an already materialized KV cache, but also by deciding which prefill-token states need to persist as full-depth cached memory in the first place.

\bibliographystyle{plainnat}
\bibliography{references}

\appendix

\section{Additional Method Details}
\label{app:method-details}

\paragraph{Training-time visibility.}
For an instruction-tuning example, let $P$ denote prefill tokens whose KV states follow shallow visibility, let $A$ denote the full-depth prefill-side anchor set, and let
$Y=\{y_1,\ldots,y_M\}$ denote assistant target tokens. During SPEED-aware supervised fine-tuning, assistant target positions follow the same visibility pattern as Decode-phase tokens under teacher forcing: they traverse all $L$ layers, while prefill tokens in $P$ remain lower-layer-only. For target token $y_t$, the visible set is
\begin{equation}
\mathcal{V}^{\mathrm{train}}_l(t)
=
\begin{cases}
P \cup A \cup Y_{<t} \cup \{y_t\}, & l \le K, \\
A \cup Y_{<t} \cup \{y_t\}, & l > K.
\end{cases}
\end{equation}
For anchor-free SPEED, $A=\emptyset$ and $P$ includes the full prefill sequence. For SPEED+BoS, $A=\{s\}$, where $s$ is the existing BoS token, and $P$ includes the remaining prefill tokens. The language-modeling objective is unchanged; the loss is computed on assistant target tokens as in standard supervised fine-tuning.

\paragraph{Post-hoc SPEED versus SPEED-aware training.}
PostHoc-SPEED applies the SPEED visibility policy only at inference time to a model trained with full upper-layer prefill-token access. This creates a train--test mismatch because upper layers were trained to rely on full-depth prefill-token KV states that are missing at inference. In contrast, SPEED-aware training exposes the model to the same prefill-token visibility constraint during fine-tuning, so upper layers learn to generate from lower-layer prompt grounding, earlier Decode-phase tokens, and the anchor set.

\paragraph{Position handling.}
Position indices are not renumbered. Prefill tokens and Decode-phase tokens keep their original positions even when prefill-token KV states are absent in upper layers. This preserves lower-layer prompt geometry and Decode-token positions. Upper-layer attention may therefore contain gaps in the visible position sequence; this is intentional because SPEED changes KV visibility, not token positional identity.

\paragraph{Anchor sets.}
The main method uses the minimal anchor set $A=\{s\}$, where $s$ is the existing BoS token. More generally, SPEED can support a small full-depth prefill-side anchor set $A$ rather than a single BoS token. We use the minimal case to isolate whether one stable prefill-side reference is sufficient to recover generation quality without restoring full upper-layer access to the prefill sequence. The BoS anchor should not be interpreted as a learned prompt summary, compressed representation, or additional memory module.

\paragraph{Decode-token cache.}
Decode-phase tokens always produce full-depth hidden states and full-depth KV states. Therefore, future Decode-phase tokens can attend to earlier Decode-phase tokens in upper layers as in Full-Attn. SPEED removes prefill-token KV states from upper-layer memory, not upper-layer Decode computation itself.

\section{Additional Experimental Setup}
\label{app:setup-details}

\paragraph{Code and data availability.}
For double-blind review, we provide two separate anonymized artifacts.
First, the NeurIPS supplemental ZIP contains the SPEED implementation, training and inference scripts, evaluation scripts, long-context efficiency measurement code, diagnostic analysis code, configuration files, dependency specifications, and reproduction instructions.
This ZIP is the code artifact and does not serve as the dataset release.
If the ZIP contains a minimal sanity-check input, it is used only to verify that the scripts execute and is not part of the experimental training or evaluation datasets.

Second, the dataset artifacts are released through an anonymized Zenodo record:
\url{https://zenodo.org/records/20057920}.
The Zenodo record contains the redistributable dataset files, documented splits, construction and filtering descriptions, checksums, metadata, and license or access notes.
Public upstream models and datasets are not redistributed unless their licenses permit redistribution; for non-redistributed assets, we provide source identifiers, download instructions, preprocessing instructions, and license or access notes.

\subsection{Existing Assets, Licenses, and Released Artifacts}
\label{app:assets}

\paragraph{Existing models, datasets, and software.}
Our experiments use publicly available models, datasets, evaluation suites, and software packages. We cite the original sources in the main paper and bibliography, and we provide an asset manifest in the anonymized supplemental ZIP that lists the version, source, license or terms of use, and redistribution policy for each asset. For public datasets and models, we do not redistribute the original assets unless their licenses permit redistribution; instead, we provide download and preprocessing instructions.

\begin{fixedtable}
\caption{
Summary of existing assets used in the experiments.
The supplemental asset manifest provides version, source, license or terms-of-use information,
and redistribution notes for each asset.
}
\label{tab:existing-assets}
\centering
\small
\setlength{\tabcolsep}{3.0pt}
\begin{tabular}{p{0.20\linewidth}p{0.33\linewidth}p{0.33\linewidth}}
\toprule
Asset type & Assets & Use in this work \\
\midrule
Base models
& Llama-3.1-8B Base and Llama-3.1-8B-Instruct
& Starting checkpoints for controlled instruction tuning and off-the-shelf compatibility experiments. \\
Instruction-tuning data
& Tulu3-style supervised fine-tuning mixture
& Controlled instruction-tuning data used for Full-IT and SPEED-aware variants. \\
General-capability evaluation
& OLMES-style TULU-3-DEV tasks, including MMLU, TruthfulQA, PopQA, BBH, DROP, GSM8K, MATH/Minerva-style math, HumanEval, HumanEval+, IFEval, and AlpacaEval-style evaluation
& Broad evaluation of instruction-tuned model quality. \\
Downstream and long-context evaluation
& HotpotQA, TriviaQA, Natural Questions, S-NIAH, and CNN/DailyMail
& Document QA, long-context retrieval, and summarization evaluation. \\
Task-adaptive training data
& Correctness-filtered teacher-generated HotpotQA data, Nemotron-Math, and OpenCodeInstruct
& Downstream adaptation for document QA, math, and code experiments. \\
Metrics and evaluation tools
& BERTScore and task-specific exact-match/F1/code/math evaluators
& Automatic evaluation of summarization, QA, code, math, and instruction-following outputs. \\
Software
& PyTorch, Hugging Face Transformers/Datasets, Accelerate, DeepSpeed, and attention/kernel libraries
& Model training, inference, evaluation, and efficiency measurement. \\
\bottomrule
\end{tabular}
\end{fixedtable}

\paragraph{New artifacts released for review.}
The anonymized supplemental ZIP and the anonymized Zenodo dataset record are intended to be complementary.
The supplemental ZIP contains code, configurations, scripts, and reproduction instructions.
The Zenodo record contains the released dataset artifacts, including redistributable derived data, documented splits, dataset-level metadata, construction and filtering notes, checksums, and license or access information.
We separate these artifacts to avoid conflating executable code with dataset release and to make the data citation persistent through Zenodo.
If accepted, both artifacts will be de-anonymized or versioned for the camera-ready release, subject to third-party license constraints.

\paragraph{Hardware.}
Table~\ref{tab:hardware-summary} summarizes the hardware used for training and evaluation. We report hardware at the level needed to reproduce the compute setting and omit run-specific scheduler, node, and job identifiers.

\begin{fixedtable}
\caption{Hardware used for training and evaluation.}
\label{tab:hardware-summary}
\centering
\begin{tabular}{p{0.34\linewidth}p{0.58\linewidth}}
\toprule
Stage & Configuration \\
\midrule
Controlled base instruction tuning & 4$\times$ NVIDIA H100 GPUs, 64 CPU cores, bf16, DeepSpeed ZeRO-3 \\
Downstream task-adaptive fine-tuning & 1$\times$ NVIDIA RTX PRO 6000-class GPU, bf16 \\
Base and downstream evaluation & 1$\times$ NVIDIA RTX PRO 6000 Blackwell-class GPU with approximately 95 GiB device memory, bf16 \\
Long-context efficiency measurement & 1$\times$ NVIDIA RTX PRO 6000 Blackwell-class GPU, batch size 1 \\
\bottomrule
\end{tabular}
\end{fixedtable}

\paragraph{Training configuration.}
Tables~\ref{tab:base-sft-config} and~\ref{tab:downstream-sft-config} report the main supervised fine-tuning and downstream task-adaptive fine-tuning configurations. Base SFT refers to the controlled instruction-tuning runs from Llama-3.1-8B Base. Downstream SFT refers to task-adaptive fine-tuning from the corresponding instruction-tuned checkpoints.

\begin{fixedtable}
\caption{Controlled base instruction-tuning configuration.}
\label{tab:base-sft-config}
\centering
\begin{tabular}{p{0.34\linewidth}p{0.58\linewidth}}
\toprule
Item & Base SFT \\
\midrule
Base model & \texttt{meta-llama/Llama-3.1-8B} \\
Dataset & \texttt{allenai/tulu-3-sft-mixture} \\
Sample count & 178{,}502 examples \\
Data preparation & Open-Instruct dataset mixer \\
Seed & 123 \\
Epochs & 2 \\
Max sequence length & 4096 \\
Learning rate / scheduler & $5\times 10^{-6}$, linear \\
Warmup ratio / weight decay & 0.03 / 0.0 \\
Precision & bf16 \\
Effective batch size & 4 GPUs $\times$ 1 sample/GPU $\times$ grad. acc. 32 = 128 \\
LoRA & Off \\
SPEED configuration & Enabled; cutoff $K$ set per run \\
Optimizations & DeepSpeed ZeRO-3, FlashAttention, gradient checkpointing \\
\bottomrule
\end{tabular}
\end{fixedtable}

\begin{fixedtable}
\caption{Downstream task-adaptive fine-tuning configuration.}
\label{tab:downstream-sft-config}
\centering
\begin{tabular}{p{0.34\linewidth}p{0.58\linewidth}}
\toprule
Item & Downstream SFT \\
\midrule
Base model & Full-IT or corresponding IT-SPEED checkpoint \\
Dataset & Task-specific adaptation data, depending on the experiment \\
Training examples & 47{,}462 DocQA examples retained from 60{,}000 candidates; task-specific counts for math and code \\
Seed & 123 \\
Epochs & 1 \\
Max sequence length & 4096 \\
Learning rate / scheduler & $5\times 10^{-6}$, linear \\
Warmup ratio / weight decay & 0.03 / 0.0 \\
Precision & bf16 \\
Effective batch size & 1 GPU $\times$ 8 samples/GPU = 8 \\
LoRA & Rank 32, alpha 64, dropout 0.05 \\
SPEED configuration & Matched to the checkpoint and cutoff used in the corresponding run \\
Optimizations & FlashAttention, gradient checkpointing \\
\bottomrule
\end{tabular}
\end{fixedtable}

\paragraph{Inference configuration.}
Tables~\ref{tab:base-eval-config} and~\ref{tab:downstream-eval-config} summarize the evaluation-time configuration. All reported evaluations use deterministic decoding.

\begin{fixedtable}
\caption{Base evaluation configuration.}
\label{tab:base-eval-config}
\centering
\begin{tabular}{p{0.34\linewidth}p{0.58\linewidth}}
\toprule
Item & Base eval \\
\midrule
Model & Full-IT or IT-SPEED checkpoint \\
Engine & Hugging Face \\
Precision / attention & bf16, SDPA \\
Batch size & 1 \\
Maximum length & 4096 \\
Decoding & temperature 0.0; \texttt{do\_sample=False} \\
Seeds & Few-shot and random-subsample seeds mostly 1234; GSM8K subsample seed 42 \\
SPEED runtime & Cutoff $K$ matched to the evaluated run; lower-only prompt prefill; causal attention; no replay \\
\bottomrule
\end{tabular}
\end{fixedtable}

\begin{fixedtable}
\caption{Downstream evaluation configuration.}
\label{tab:downstream-eval-config}
\centering
\begin{tabular}{p{0.34\linewidth}p{0.58\linewidth}}
\toprule
Item & Downstream eval \\
\midrule
Model & Downstream-adapted checkpoint or off-the-shelf instruction checkpoint for the pilot \\
Engine & Hugging Face \\
Precision / attention & bf16, SDPA \\
Batching & Per-sample loop \\
Maximum length & \texttt{max\_doc\_tokens}=130000; \texttt{max\_new\_tokens}=1024 \\
Decoding & temperature 0.0; top-$p$ 1.0; \texttt{do\_sample=False} \\
Seeds & Few-shot seed 1234; NQ and S-NIAH shuffle seeds 42 \\
SPEED runtime & Matched to the evaluated checkpoint and reported method \\
\bottomrule
\end{tabular}
\end{fixedtable}

\paragraph{Evaluation datasets.}
Tables~\ref{tab:tulu-eval-datasets} and~\ref{tab:downstream-eval-datasets} report the benchmark splits, number of shots, and sample counts used in evaluation.

\begin{fixedtable}
\caption{TULU-3-DEV evaluation datasets.}
\label{tab:tulu-eval-datasets}
\centering
\begin{tabular}{p{0.36\linewidth}p{0.18\linewidth}p{0.16\linewidth}p{0.16\linewidth}}
\toprule
Dataset / task & Split & Shots & Samples \\
\midrule
GSM8K & test & 8 & 1{,}319 \\
BBH aggregate & test & 3 & 6{,}511 \\
DROP \texttt{drop::llama3} & validation & 3 & 9{,}536 \\
DROP \texttt{drop:chat} & validation & 3 & 9{,}536 \\
Minerva Math & test & 4 & 5{,}000 \\
HumanEval & test & 0 & 164 \\
HumanEval+ & test & 0 & 164 \\
IFEval & train & 0 & 541 \\
PopQA & test & 15 & 14{,}267 \\
MMLU MC & test & 5 & 14{,}042 \\
MMLU CoT & test & 0 & 14{,}042 \\
AlpacaEval v3 & test & 0 & 805 \\
TruthfulQA MC & validation & 6 & 817 \\
\bottomrule
\end{tabular}
\end{fixedtable}

\begin{fixedtable}
\caption{Downstream evaluation datasets.}
\label{tab:downstream-eval-datasets}
\centering
\begin{tabular}{p{0.34\linewidth}p{0.22\linewidth}p{0.16\linewidth}p{0.16\linewidth}}
\toprule
Dataset & Split & Shots & Samples \\
\midrule
HotpotQA & validation/dev & 3 & 1{,}000 \\
Natural Questions & validation/dev & 3 & 1{,}000 \\
TriviaQA & validation/dev & 3 & 1{,}000 \\
S-NIAH & validation/dev & 0 & 1{,}000 \\
CNN/Dailymail & validation/dev & 3 & 1{,}000 \\
MathBench & test & 3 & 1{,}000 \\
BigCodeBench & v0.1.4 & 3 & 1{,}000 \\
\bottomrule
\end{tabular}
\end{fixedtable}

\subsection{Layer-wise Diagnostic Configuration}
\label{app:layer-diagnostic-setup}

The layer-wise diagnostic reuses TULU-3-DEV request prompts and does not constitute a separate benchmark evaluation. We run Full-IT with greedy decoding and collect layer-wise attention, conditional prompt entropy, and hidden-state trajectory statistics during Decode. Prefill is executed only to construct the KV cache; the reported attention quantities measure where generated-token queries attend during Decode.

We group tasks into five categories. Math includes \texttt{gsm8k::tulu} and \texttt{minerva\_math::tulu}; Coding includes \texttt{codex\_humaneval::tulu} and \texttt{codex\_humanevalplus::tulu}; Reasoning includes \texttt{bbh:cot-v1::tulu} and \texttt{drop:chat}; Knowledge includes \texttt{mmlu:mc::tulu}, \texttt{truthfulqa::tulu}, and \texttt{popqa::tulu}; Instruction includes \texttt{ifeval::tulu} and \texttt{alpaca\_eval\_v3::tulu}. We use 100 examples from each root task, resulting in 200 examples for Math, Coding, Reasoning, and Instruction, and 300 examples for Knowledge.

For each layer $l$, we compute attention mass from the current Decode-phase token to user-prompt tokens, BoS, and earlier Decode-phase tokens:
\begin{equation}
A_l^U,\quad A_l^{\mathrm{BoS}},\quad A_l^{D<t}.
\end{equation}
The user-prompt span includes only the content of messages with role \texttt{user} under the chat template; BoS, system text, role headers, and assistant headers are excluded from $U$.

We also compute conditional prompt entropy by renormalizing attention over user-prompt tokens:
\begin{equation}
H_l^U
=
-\sum_{i\in U}
\tilde{a}_{l,i}\log \tilde{a}_{l,i},
\qquad
\tilde{a}_{l,i}
=
\frac{a_{l,i}}{\sum_{j\in U} a_{l,j}}.
\end{equation}
Lower $H_l^U$ indicates more selective prompt access within the user prompt. Finally, we compute all-token hidden-trajectory straightening from the reduction in average token-position trajectory curvature relative to layer 1. We use these statistics as category-level diagnostics, not as causal proofs or per-example cutoff predictors.

\section{Full General-Capability Results}
\label{app:full-results}

\begin{fixedtable}
\caption{
Full general capability results after Tulu-style instruction tuning.
The table reports aggregate score, knowledge, reasoning, code, math, and instruction-following benchmarks.
}
\label{tab:general-full}
\centering
\small
\setlength{\tabcolsep}{2.0pt}
\begin{tabular}{@{}lcccccccccccc@{}}
\toprule
\multirow{2}{*}{Method}
& \multirow{2}{*}{Avg.}
& \multicolumn{3}{c}{Knowledge}
& \multicolumn{2}{c}{Reasoning}
& \multicolumn{2}{c}{Code}
& \multicolumn{2}{c}{Math}
& \multicolumn{2}{c}{Instruction} \\
\cmidrule(lr){3-5}
\cmidrule(lr){6-7}
\cmidrule(lr){8-9}
\cmidrule(lr){10-11}
\cmidrule(l){12-13}
& & MMLU & TQA & PopQA & BBH & DROP & CHE & CHE+ & GSM & MATH & IFEval & AE2 \\
\midrule
Full-IT ($K=32$) & 51.4 & 59.7 & 46.9 & 28.1 & 63.9 & 51.7 & 78.5 & 70.6 & 67.2 & 25.9 & 63.2 & 9.0 \\
IT-SPEED-28+BoS & 51.3 & 58.9 & 50.3 & 27.8 & 63.8 & 50.3 & 78.7 & 71.7 & 68.2 & 24.9 & 62.5 & 7.3 \\
IT-SPEED-28 & 50.2 & 59.8 & 50.1 & 22.0 & 64.1 & 48.1 & 78.0 & 69.7 & 66.9 & 23.6 & 63.4 & 6.6 \\
IT-SPEED-24+BoS & 51.2 & 59.6 & 51.2 & 27.2 & 62.7 & 53.2 & 79.0 & 71.7 & 66.0 & 24.6 & 60.6 & 7.2 \\
IT-SPEED-24 & 49.1 & 57.4 & 49.4 & 18.3 & 59.7 & 49.9 & 79.6 & 71.6 & 64.1 & 22.1 & 60.4 & 8.2 \\
IT-SPEED-20+BoS & 49.9 & 58.7 & 50.5 & 26.2 & 61.4 & 50.7 & 81.1 & 69.8 & 64.1 & 21.7 & 56.9 & 7.7 \\
IT-SPEED-20 & 48.6 & 57.4 & 49.4 & 20.0 & 62.3 & 48.8 & 78.1 & 69.1 & 64.4 & 21.2 & 57.7 & 6.3 \\
IT-SPEED-16+BoS & 45.4 & 55.9 & 50.2 & 22.9 & 52.0 & 35.9 & 74.8 & 67.1 & 62.4 & 18.7 & 52.7 & 7.0 \\
IT-SPEED-16 & 44.3 & 51.7 & 49.8 & 16.4 & 53.5 & 37.6 & 78.9 & 69.2 & 57.8 & 14.3 & 51.4 & 6.8 \\
\bottomrule
\end{tabular}
\end{fixedtable}

\section{Full Efficiency Measurements}
\label{app:efficiency-full}

All efficiency measurements use Llama-3.1-8B with a fixed continuation length of 128 generated tokens.
We vary the prompt length from 1K to 128K tokens and report the mean and standard deviation over five repeats.
TTFT is measured before the first generated token, and TPOT is the average latency per generated token over the fixed continuation.
Active KV-cache memory counts materialized KV tensors, including the fixed 128-token continuation cache.
Estimated total FLOPs are computed from the layer-token computation proxy used in our cost analysis and are intended for relative comparison under the same measurement protocol.

For readability, we separate the SPEED cutoff sweep from the stage-aware $K=24$ comparison.
Each latency cell reports the raw value on the first line and the speedup relative to Full-Attn on the second line.
Each memory or FLOPs cell reports the raw value on the first line and the reduction relative to Full-Attn on the second line.

\subsection{SPEED cutoff sweep}

\begin{fixedtable}
\caption{TTFT in milliseconds across prompt lengths for the SPEED cutoff sweep. The second line reports speedup relative to Full-Attn.}
\label{tab:ttft-speed-sweep}
\centering
\small
\setlength{\tabcolsep}{3.0pt}
\begin{tabular}{@{}lccccc@{}}
\toprule
Prompt & Full-Attn & SPEED-16 & SPEED-20 & SPEED-24 & SPEED-28 \\
\midrule
1K &
\metriccell{$77.64 \pm 0.09$}{$1.00\times$} &
\metriccell{$47.45 \pm 0.07$}{$1.64\times$} &
\metriccell{$55.39 \pm 0.15$}{$1.40\times$} &
\metriccell{$63.29 \pm 0.04$}{$1.23\times$} &
\metriccell{$70.96 \pm 0.03$}{$1.09\times$} \\
2K &
\metriccell{$118.34 \pm 0.10$}{$1.00\times$} &
\metriccell{$67.93 \pm 0.03$}{$1.74\times$} &
\metriccell{$80.85 \pm 0.05$}{$1.46\times$} &
\metriccell{$93.95 \pm 0.12$}{$1.26\times$} &
\metriccell{$107.00 \pm 0.12$}{$1.11\times$} \\
4K &
\metriccell{$237.35 \pm 0.31$}{$1.00\times$} &
\metriccell{$119.95 \pm 0.14$}{$1.98\times$} &
\metriccell{$149.50 \pm 0.23$}{$1.59\times$} &
\metriccell{$180.80 \pm 0.39$}{$1.31\times$} &
\metriccell{$211.95 \pm 0.32$}{$1.12\times$} \\
8K &
\metriccell{$530.77 \pm 0.80$}{$1.00\times$} &
\metriccell{$267.36 \pm 0.18$}{$1.99\times$} &
\metriccell{$335.35 \pm 0.16$}{$1.58\times$} &
\metriccell{$404.41 \pm 0.41$}{$1.31\times$} &
\metriccell{$473.45 \pm 0.48$}{$1.12\times$} \\
16K &
\metriccell{$1206.27 \pm 2.98$}{$1.00\times$} &
\metriccell{$607.59 \pm 0.57$}{$1.99\times$} &
\metriccell{$762.76 \pm 1.12$}{$1.58\times$} &
\metriccell{$918.59 \pm 1.20$}{$1.31\times$} &
\metriccell{$1075.02 \pm 0.96$}{$1.12\times$} \\
32K &
\metriccell{$2896.03 \pm 14.67$}{$1.00\times$} &
\metriccell{$1455.34 \pm 2.19$}{$1.99\times$} &
\metriccell{$1827.23 \pm 4.88$}{$1.58\times$} &
\metriccell{$2197.52 \pm 6.30$}{$1.32\times$} &
\metriccell{$2570.24 \pm 7.90$}{$1.13\times$} \\
64K &
\metriccell{$7674.12 \pm 43.93$}{$1.00\times$} &
\metriccell{$3833.26 \pm 14.42$}{$2.00\times$} &
\metriccell{$4818.95 \pm 18.20$}{$1.59\times$} &
\metriccell{$5788.21 \pm 21.45$}{$1.33\times$} &
\metriccell{$6783.49 \pm 25.91$}{$1.13\times$} \\
128K &
\metriccell{$22898.60 \pm 89.18$}{$1.00\times$} &
\metriccell{$11410.25 \pm 44.32$}{$2.01\times$} &
\metriccell{$14320.81 \pm 37.49$}{$1.60\times$} &
\metriccell{$17199.90 \pm 38.34$}{$1.33\times$} &
\metriccell{$20130.24 \pm 34.82$}{$1.14\times$} \\
\bottomrule
\end{tabular}
\end{fixedtable}

\begin{fixedtable}
\caption{TPOT in milliseconds per generated token across prompt lengths for the SPEED cutoff sweep. The second line reports speedup relative to Full-Attn.}
\label{tab:tpot-speed-sweep}
\centering
\small
\setlength{\tabcolsep}{3.0pt}
\begin{tabular}{@{}lccccc@{}}
\toprule
Prompt & Full-Attn & SPEED-16 & SPEED-20 & SPEED-24 & SPEED-28 \\
\midrule
1K &
\metriccell{$24.05 \pm 0.06$}{$1.00\times$} &
\metriccell{$24.06 \pm 0.02$}{$1.00\times$} &
\metriccell{$24.17 \pm 0.04$}{$1.00\times$} &
\metriccell{$24.15 \pm 0.04$}{$1.00\times$} &
\metriccell{$24.17 \pm 0.11$}{$1.00\times$} \\
2K &
\metriccell{$24.22 \pm 0.27$}{$1.00\times$} &
\metriccell{$24.16 \pm 0.13$}{$1.00\times$} &
\metriccell{$24.20 \pm 0.04$}{$1.00\times$} &
\metriccell{$24.11 \pm 0.01$}{$1.00\times$} &
\metriccell{$24.19 \pm 0.04$}{$1.00\times$} \\
4K &
\metriccell{$24.32 \pm 0.17$}{$1.00\times$} &
\metriccell{$24.32 \pm 0.34$}{$1.00\times$} &
\metriccell{$24.18 \pm 0.02$}{$1.01\times$} &
\metriccell{$24.15 \pm 0.03$}{$1.01\times$} &
\metriccell{$24.18 \pm 0.03$}{$1.01\times$} \\
8K &
\metriccell{$24.23 \pm 0.17$}{$1.00\times$} &
\metriccell{$24.17 \pm 0.05$}{$1.00\times$} &
\metriccell{$24.13 \pm 0.01$}{$1.00\times$} &
\metriccell{$24.14 \pm 0.04$}{$1.00\times$} &
\metriccell{$24.18 \pm 0.01$}{$1.00\times$} \\
16K &
\metriccell{$23.97 \pm 0.01$}{$1.00\times$} &
\metriccell{$24.02 \pm 0.08$}{$1.00\times$} &
\metriccell{$24.16 \pm 0.01$}{$0.99\times$} &
\metriccell{$24.16 \pm 0.02$}{$0.99\times$} &
\metriccell{$24.20 \pm 0.01$}{$0.99\times$} \\
32K &
\metriccell{$24.03 \pm 0.05$}{$1.00\times$} &
\metriccell{$23.89 \pm 0.14$}{$1.01\times$} &
\metriccell{$24.15 \pm 0.02$}{$1.00\times$} &
\metriccell{$24.12 \pm 0.09$}{$1.00\times$} &
\metriccell{$24.30 \pm 0.24$}{$0.99\times$} \\
64K &
\metriccell{$29.47 \pm 0.01$}{$1.00\times$} &
\metriccell{$24.03 \pm 0.07$}{$1.23\times$} &
\metriccell{$24.15 \pm 0.04$}{$1.22\times$} &
\metriccell{$25.52 \pm 0.02$}{$1.15\times$} &
\metriccell{$27.54 \pm 0.01$}{$1.07\times$} \\
128K &
\metriccell{$46.96 \pm 0.00$}{$1.00\times$} &
\metriccell{$30.28 \pm 0.01$}{$1.55\times$} &
\metriccell{$34.46 \pm 0.01$}{$1.36\times$} &
\metriccell{$38.64 \pm 0.00$}{$1.22\times$} &
\metriccell{$42.80 \pm 0.00$}{$1.10\times$} \\
\bottomrule
\end{tabular}
\end{fixedtable}

\begin{fixedtable}
\caption{
Active KV-cache memory in GiB for the SPEED cutoff sweep.
The second line reports memory reduction relative to Full-Attn.
}
\label{tab:active-kv-speed-sweep}
\centering
\small
\setlength{\tabcolsep}{3.0pt}
\begin{tabular}{@{}lccccc@{}}
\toprule
Prompt & Full-Attn & SPEED-16 & SPEED-20 & SPEED-24 & SPEED-28 \\
\midrule
1K &
\metriccell{0.141}{0.0\%} &
\metriccell{0.078}{44.7\%} &
\metriccell{0.094}{33.3\%} &
\metriccell{0.109}{22.7\%} &
\metriccell{0.125}{11.3\%} \\
2K &
\metriccell{0.266}{0.0\%} &
\metriccell{0.141}{47.0\%} &
\metriccell{0.172}{35.3\%} &
\metriccell{0.203}{23.7\%} &
\metriccell{0.234}{12.0\%} \\
4K &
\metriccell{0.516}{0.0\%} &
\metriccell{0.266}{48.4\%} &
\metriccell{0.328}{36.4\%} &
\metriccell{0.391}{24.2\%} &
\metriccell{0.453}{12.2\%} \\
8K &
\metriccell{1.016}{0.0\%} &
\metriccell{0.516}{49.2\%} &
\metriccell{0.641}{36.9\%} &
\metriccell{0.766}{24.6\%} &
\metriccell{0.891}{12.3\%} \\
16K &
\metriccell{2.016}{0.0\%} &
\metriccell{1.016}{49.6\%} &
\metriccell{1.266}{37.2\%} &
\metriccell{1.516}{24.8\%} &
\metriccell{1.766}{12.4\%} \\
32K &
\metriccell{4.016}{0.0\%} &
\metriccell{2.016}{49.8\%} &
\metriccell{2.516}{37.4\%} &
\metriccell{3.016}{24.9\%} &
\metriccell{3.516}{12.5\%} \\
64K &
\metriccell{8.016}{0.0\%} &
\metriccell{4.016}{49.9\%} &
\metriccell{5.016}{37.4\%} &
\metriccell{6.016}{25.0\%} &
\metriccell{7.016}{12.5\%} \\
128K &
\metriccell{16.016}{0.0\%} &
\metriccell{8.016}{50.0\%} &
\metriccell{10.016}{37.5\%} &
\metriccell{12.016}{25.0\%} &
\metriccell{14.016}{12.5\%} \\
\bottomrule
\end{tabular}
\end{fixedtable}

\begin{fixedtable}
\caption{
Estimated total FLOPs in teraFLOPs for the SPEED cutoff sweep.
The second line reports FLOPs reduction relative to Full-Attn.
}
\label{tab:flops-speed-sweep}
\centering
\small
\setlength{\tabcolsep}{3.0pt}
\begin{tabular}{@{}lccccc@{}}
\toprule
Prompt & Full-Attn & SPEED-16 & SPEED-20 & SPEED-24 & SPEED-28 \\
\midrule
1K &
\metriccell{16.828}{0.0\%} &
\metriccell{9.432}{44.0\%} &
\metriccell{11.281}{33.0\%} &
\metriccell{13.130}{22.0\%} &
\metriccell{14.979}{11.0\%} \\
2K &
\metriccell{32.853}{0.0\%} &
\metriccell{17.479}{46.8\%} &
\metriccell{21.323}{35.1\%} &
\metriccell{25.166}{23.4\%} &
\metriccell{29.010}{11.7\%} \\
4K &
\metriccell{68.228}{0.0\%} &
\metriccell{35.235}{48.4\%} &
\metriccell{43.483}{36.3\%} &
\metriccell{51.731}{24.2\%} &
\metriccell{59.980}{12.1\%} \\
8K &
\metriccell{152.274}{0.0\%} &
\metriccell{77.395}{49.2\%} &
\metriccell{96.115}{36.9\%} &
\metriccell{114.834}{24.6\%} &
\metriccell{133.554}{12.3\%} \\
16K &
\metriccell{373.554}{0.0\%} &
\metriccell{188.310}{49.6\%} &
\metriccell{234.621}{37.2\%} &
\metriccell{280.932}{24.8\%} &
\metriccell{327.243}{12.4\%} \\
32K &
\metriccell{1028.870}{0.0\%} &
\metriccell{516.518}{49.8\%} &
\metriccell{644.606}{37.3\%} &
\metriccell{772.694}{24.9\%} &
\metriccell{900.782}{12.4\%} \\
64K &
\metriccell{3190.525}{0.0\%} &
\metriccell{1598.445}{49.9\%} &
\metriccell{1996.465}{37.4\%} &
\metriccell{2394.485}{25.0\%} &
\metriccell{2792.505}{12.5\%} \\
128K &
\metriccell{10917.923}{0.0\%} &
\metriccell{5464.343}{50.0\%} &
\metriccell{6827.738}{37.5\%} &
\metriccell{8191.133}{25.0\%} &
\metriccell{9554.528}{12.5\%} \\
\bottomrule
\end{tabular}
\end{fixedtable}

\subsection{Stage-aware Prefill baselines at K=24}

\begin{fixedtable}
\caption{TTFT in milliseconds for stage-aware Prefill baselines at $K=24$. The second line reports speedup relative to Full-Attn.}
\label{tab:ttft-stage-aware}
\centering
\small
\setlength{\tabcolsep}{3.0pt}
\begin{tabular}{@{}lcccc@{}}
\toprule
Prompt & Full-Attn & SwiftKV-24 & POP-24 & SPEED-24 \\
\midrule
1K &
\metriccell{$77.64 \pm 0.09$}{$1.00\times$} &
\metriccell{$74.36 \pm 3.78$}{$1.04\times$} &
\metriccell{$72.67 \pm 0.36$}{$1.07\times$} &
\metriccell{$63.29 \pm 0.04$}{$1.23\times$} \\
2K &
\metriccell{$118.34 \pm 0.10$}{$1.00\times$} &
\metriccell{$102.26 \pm 1.37$}{$1.16\times$} &
\metriccell{$107.42 \pm 7.80$}{$1.10\times$} &
\metriccell{$93.95 \pm 0.12$}{$1.26\times$} \\
4K &
\metriccell{$237.35 \pm 0.31$}{$1.00\times$} &
\metriccell{$189.22 \pm 1.67$}{$1.25\times$} &
\metriccell{$187.92 \pm 0.75$}{$1.26\times$} &
\metriccell{$180.80 \pm 0.39$}{$1.31\times$} \\
8K &
\metriccell{$530.77 \pm 0.80$}{$1.00\times$} &
\metriccell{$398.67 \pm 2.03$}{$1.33\times$} &
\metriccell{$407.13 \pm 0.19$}{$1.30\times$} &
\metriccell{$404.41 \pm 0.41$}{$1.31\times$} \\
16K &
\metriccell{$1206.27 \pm 2.98$}{$1.00\times$} &
\metriccell{$899.59 \pm 8.64$}{$1.34\times$} &
\metriccell{$921.62 \pm 6.07$}{$1.31\times$} &
\metriccell{$918.59 \pm 1.20$}{$1.31\times$} \\
32K &
\metriccell{$2896.03 \pm 14.67$}{$1.00\times$} &
\metriccell{$2145.10 \pm 10.61$}{$1.35\times$} &
\metriccell{$2182.87 \pm 7.45$}{$1.33\times$} &
\metriccell{$2197.52 \pm 6.30$}{$1.32\times$} \\
64K &
\metriccell{$7674.12 \pm 43.93$}{$1.00\times$} &
\metriccell{$5690.00 \pm 32.19$}{$1.35\times$} &
\metriccell{$5757.18 \pm 21.67$}{$1.33\times$} &
\metriccell{$5788.21 \pm 21.45$}{$1.33\times$} \\
128K &
\metriccell{$22898.60 \pm 89.18$}{$1.00\times$} &
\metriccell{$17019.02 \pm 55.18$}{$1.35\times$} &
\metriccell{$17093.09 \pm 47.74$}{$1.34\times$} &
\metriccell{$17199.90 \pm 38.34$}{$1.33\times$} \\
\bottomrule
\end{tabular}
\end{fixedtable}

\begin{fixedtable}
\caption{TPOT in milliseconds per generated token for stage-aware Prefill baselines at $K=24$. The second line reports speedup relative to Full-Attn.}
\label{tab:tpot-stage-aware}
\centering
\small
\setlength{\tabcolsep}{3.0pt}
\begin{tabular}{@{}lcccc@{}}
\toprule
Prompt & Full-Attn & SwiftKV-24 & POP-24 & SPEED-24 \\
\midrule
1K &
\metriccell{$24.05 \pm 0.06$}{$1.00\times$} &
\metriccell{$26.38 \pm 0.45$}{$0.91\times$} &
\metriccell{$24.24 \pm 0.03$}{$0.99\times$} &
\metriccell{$24.15 \pm 0.04$}{$1.00\times$} \\
2K &
\metriccell{$24.22 \pm 0.27$}{$1.00\times$} &
\metriccell{$25.83 \pm 0.60$}{$0.94\times$} &
\metriccell{$24.24 \pm 0.09$}{$1.00\times$} &
\metriccell{$24.11 \pm 0.01$}{$1.00\times$} \\
4K &
\metriccell{$24.32 \pm 0.17$}{$1.00\times$} &
\metriccell{$25.65 \pm 0.11$}{$0.95\times$} &
\metriccell{$24.65 \pm 0.22$}{$0.99\times$} &
\metriccell{$24.15 \pm 0.03$}{$1.01\times$} \\
8K &
\metriccell{$24.23 \pm 0.17$}{$1.00\times$} &
\metriccell{$25.52 \pm 0.24$}{$0.95\times$} &
\metriccell{$25.34 \pm 0.75$}{$0.96\times$} &
\metriccell{$24.14 \pm 0.04$}{$1.00\times$} \\
16K &
\metriccell{$23.97 \pm 0.01$}{$1.00\times$} &
\metriccell{$25.69 \pm 0.17$}{$0.93\times$} &
\metriccell{$26.76 \pm 0.24$}{$0.90\times$} &
\metriccell{$24.16 \pm 0.02$}{$0.99\times$} \\
32K &
\metriccell{$24.03 \pm 0.05$}{$1.00\times$} &
\metriccell{$25.59 \pm 0.14$}{$0.94\times$} &
\metriccell{$26.55 \pm 0.09$}{$0.91\times$} &
\metriccell{$24.12 \pm 0.09$}{$1.00\times$} \\
64K &
\metriccell{$29.47 \pm 0.01$}{$1.00\times$} &
\metriccell{$30.02 \pm 0.11$}{$0.98\times$} &
\metriccell{$30.50 \pm 0.06$}{$0.97\times$} &
\metriccell{$25.52 \pm 0.02$}{$1.15\times$} \\
128K &
\metriccell{$46.96 \pm 0.00$}{$1.00\times$} &
\metriccell{$47.35 \pm 0.09$}{$0.99\times$} &
\metriccell{$47.75 \pm 0.05$}{$0.98\times$} &
\metriccell{$38.64 \pm 0.00$}{$1.22\times$} \\
\bottomrule
\end{tabular}
\end{fixedtable}

\begin{fixedtable}
\caption{
Active KV-cache memory in GiB for stage-aware Prefill baselines at $K=24$.
The second line reports memory reduction relative to Full-Attn.
}
\label{tab:active-kv-stage-aware}
\centering
\small
\setlength{\tabcolsep}{3.0pt}
\begin{tabular}{@{}lcccc@{}}
\toprule
Prompt & Full-Attn & SwiftKV-24 & POP-24 & SPEED-24 \\
\midrule
1K &
\metriccell{0.141}{0.0\%} &
\metriccell{0.123}{12.8\%} &
\metriccell{0.141}{0.0\%} &
\metriccell{0.109}{22.7\%} \\
2K &
\metriccell{0.266}{0.0\%} &
\metriccell{0.232}{12.8\%} &
\metriccell{0.266}{0.0\%} &
\metriccell{0.203}{23.7\%} \\
4K &
\metriccell{0.516}{0.0\%} &
\metriccell{0.451}{12.6\%} &
\metriccell{0.516}{0.0\%} &
\metriccell{0.391}{24.2\%} \\
8K &
\metriccell{1.016}{0.0\%} &
\metriccell{0.889}{12.5\%} &
\metriccell{1.016}{0.0\%} &
\metriccell{0.766}{24.6\%} \\
16K &
\metriccell{2.016}{0.0\%} &
\metriccell{1.764}{12.5\%} &
\metriccell{2.016}{0.0\%} &
\metriccell{1.516}{24.8\%} \\
32K &
\metriccell{4.016}{0.0\%} &
\metriccell{3.514}{12.5\%} &
\metriccell{4.016}{0.0\%} &
\metriccell{3.016}{24.9\%} \\
64K &
\metriccell{8.016}{0.0\%} &
\metriccell{7.014}{12.5\%} &
\metriccell{8.016}{0.0\%} &
\metriccell{6.016}{25.0\%} \\
128K &
\metriccell{16.016}{0.0\%} &
\metriccell{14.014}{12.5\%} &
\metriccell{16.016}{0.0\%} &
\metriccell{12.016}{25.0\%} \\
\bottomrule
\end{tabular}
\end{fixedtable}

\begin{fixedtable}
\caption{
Estimated total FLOPs in teraFLOPs for stage-aware Prefill baselines at $K=24$.
The second line reports FLOPs reduction relative to Full-Attn.
}
\label{tab:flops-stage-aware}
\centering
\small
\setlength{\tabcolsep}{3.0pt}
\begin{tabular}{@{}lcccc@{}}
\toprule
Prompt & Full-Attn & SwiftKV-24 & POP-24 & SPEED-24 \\
\midrule
1K &
\metriccell{16.828}{0.0\%} &
\metriccell{13.180}{21.7\%} &
\metriccell{13.257}{21.2\%} &
\metriccell{13.130}{22.0\%} \\
2K &
\metriccell{32.853}{0.0\%} &
\metriccell{25.284}{23.0\%} &
\metriccell{25.430}{22.6\%} &
\metriccell{25.166}{23.4\%} \\
4K &
\metriccell{68.228}{0.0\%} &
\metriccell{51.986}{23.8\%} &
\metriccell{52.269}{23.4\%} &
\metriccell{51.731}{24.2\%} \\
8K &
\metriccell{152.274}{0.0\%} &
\metriccell{115.363}{24.2\%} &
\metriccell{115.921}{23.9\%} &
\metriccell{114.834}{24.6\%} \\
16K &
\metriccell{373.554}{0.0\%} &
\metriccell{282.008}{24.5\%} &
\metriccell{283.116}{24.2\%} &
\metriccell{280.932}{24.8\%} \\
32K &
\metriccell{1028.870}{0.0\%} &
\metriccell{774.866}{24.7\%} &
\metriccell{777.073}{24.5\%} &
\metriccell{772.694}{24.9\%} \\
64K &
\metriccell{3190.525}{0.0\%} &
\metriccell{2398.847}{24.8\%} &
\metriccell{2403.253}{24.7\%} &
\metriccell{2394.485}{25.0\%} \\
128K &
\metriccell{10917.923}{0.0\%} &
\metriccell{8199.875}{24.9\%} &
\metriccell{8208.680}{24.8\%} &
\metriccell{8191.133}{25.0\%} \\
\bottomrule
\end{tabular}
\end{fixedtable}

\paragraph{Summary.}
At 128K context, SPEED-24 reaches $1.33\times$ TTFT speedup and $1.22\times$ TPOT speedup over Full-Attn, while reducing active KV memory from 16.016 GiB to 12.016 GiB.
POP-24 closely matches SPEED-24 in Prefill-side cost, with similar estimated total FLOPs (8208.680T for POP-24 vs. 8191.133T for SPEED-24) and similar TTFT speedup, but retains the full active KV footprint and shows no TPOT improvement under this protocol.
SwiftKV-24 also reaches a similar TTFT speedup and reduces active KV memory by 12.5\%, but does not improve TPOT under our measurement.
Thus, the stage-aware comparison separates Prefill-side acceleration from Decode-time memory-interface changes: SPEED reduces repeated upper-layer prefill-token attention and active KV memory by removing the long prefill sequence from upper-layer Decode visibility.

\section{Layer-wise Diagnostics}
\label{app:layer-diagnostics}

\begin{figure}[!htbp]
    \centering
    \includegraphics[width=0.96\linewidth]{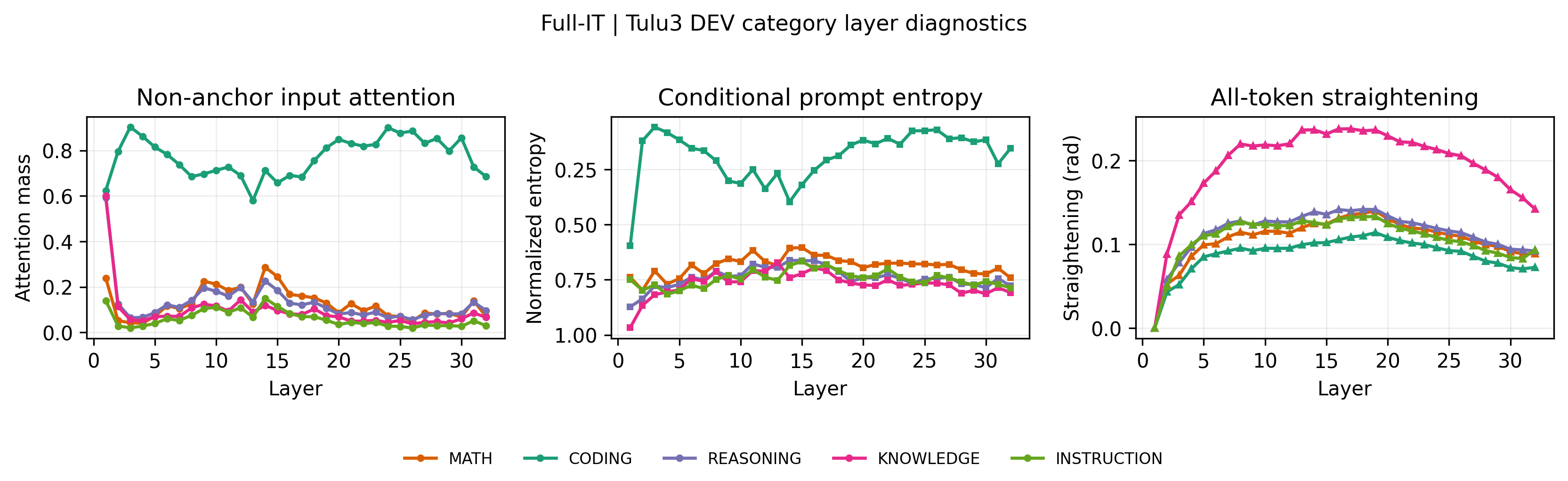}
    \caption{
    Category-level layer diagnostics on Full-IT using TULU-3-DEV request prompts.
    Left: Decode-token attention mass to user-prompt tokens.
    Middle: normalized conditional prompt entropy over user-prompt tokens; the y-axis is inverted, so higher curves indicate lower entropy and more selective prompt access.
    Right: all-token hidden-trajectory straightening.
    Across most non-coding categories, selective prompt access tends to occur before the later straightening peak; Coding shows earlier prompt-selectivity timing.
    }
    \label{fig:tulu3dev-layer-diagnostics}
\end{figure}

Figure~\ref{fig:tulu3dev-layer-diagnostics} visualizes the category-level layer diagnostics used to interpret cutoff behavior. This analysis reuses TULU-3-DEV request prompts and measures layer-wise behavior during Decode in Full-IT. The left panel reports Decode-token attention mass to user-prompt tokens. The middle panel reports conditional prompt entropy, computed by renormalizing attention over user-prompt tokens; lower entropy indicates more selective prompt access. The entropy axis is inverted so that upward movement corresponds to stronger prompt selectivity. The right panel reports all-token hidden-trajectory straightening, following trajectory-straightening analyses of predictive representations~\citep{henaff2021primary,hosseini2023large}.

Table~\ref{tab:layer-diagnostics-full} summarizes the peak layers extracted from these curves. Layer indices are 1-based. Prompt peak denotes the layer with maximum Decode-token attention mass to user-prompt tokens. BoS peak and Decode-token peak denote the corresponding peaks for BoS and earlier Decode-phase tokens. Entropy min is the layer with minimum conditional prompt entropy. Straight. peak is the layer with maximum all-token hidden-trajectory straightening. Ent.-Str. reports the difference between the entropy-minimum layer and the straightening-peak layer; negative values indicate that selective prompt access occurs before the straightening peak. Corr. is the layer-wise correlation between straightening and prompt selectivity.

\begin{fixedtable}
\caption{
Full layer-wise diagnostics on Full-IT using TULU-3-DEV prompts.
Layer indices are 1-based.
Ent.-Str. is entropy min minus straightening peak; Corr. is the layer-wise correlation between straightening and prompt selectivity.
}
\label{tab:layer-diagnostics-full}
\centering
\small
\setlength{\tabcolsep}{3.0pt}
\begin{tabular}{lcccccccc}
\toprule
Category & $n$ & Prompt peak & BoS peak & Decode-token peak & Entropy min & Straight. peak & Ent.-Str. & Corr. \\
\midrule
Math & 200 & L14 & L3 & L13 & L15 & L19 & $-4$ & 0.670 \\
Coding & 200 & L3 & L3 & L13 & L3 & L19 & $-16$ & 0.251 \\
Reasoning & 200 & L1 & L24 & L13 & L14 & L18 & $-4$ & 0.843 \\
Knowledge & 300 & L1 & L4 & L13 & L13 & L17 & $-4$ & 0.898 \\
Instruction & 200 & L14 & L3 & L1 & L15 & L19 & $-4$ & 0.441 \\
\bottomrule
\end{tabular}
\end{fixedtable}

Table~\ref{tab:layer-diagnostics-sample-alignment} reports sample-level alignment statistics between the conditional-entropy minimum and the straightening peak. For each example, we compute
$\Delta = l_{\mathrm{entropy\;min}} - l_{\mathrm{straightening\;peak}}$.
Negative values indicate that the entropy minimum occurs before the straightening peak. Exact, Within-1, and Within-2 report the fraction of examples where the two layers are exactly aligned or within one/two layers.

\begin{fixedtable}
\caption{
Sample-level alignment between conditional prompt-entropy minima and straightening peaks.
$\Delta$ is entropy-min layer minus straightening-peak layer.
}
\label{tab:layer-diagnostics-sample-alignment}
\centering
\small
\setlength{\tabcolsep}{4.0pt}
\begin{tabular}{lccccc}
\toprule
Category & $\Delta$ mean & $\Delta$ std. & Exact & Within-1 & Within-2 \\
\midrule
Math & $-4.94$ & 1.69 & 0.000 & 0.000 & 0.010 \\
Coding & $-12.63$ & 7.49 & 0.000 & 0.000 & 0.040 \\
Reasoning & $-2.02$ & 2.24 & 0.205 & 0.295 & 0.660 \\
Knowledge & $-2.27$ & 4.50 & 0.143 & 0.160 & 0.183 \\
Instruction & $-0.38$ & 8.55 & 0.030 & 0.095 & 0.185 \\
\bottomrule
\end{tabular}
\end{fixedtable}

The figure and tables reveal a task-dependent structure. First, raw prompt-attention mass alone can be misleading. Reasoning and Knowledge have prompt-mass peaks at L1, but their conditional prompt-entropy minima occur much later, at L14 and L13. This suggests that early layers may attend broadly to the prompt, while selective prompt access emerges later. Math and Instruction show a clearer access-to-stabilization ordering: prompt mass peaks at L14, conditional prompt entropy reaches its minimum at L15, and straightening peaks at L19.

Coding differs from the other categories. Its prompt peak and conditional-entropy minimum both occur at L3, whereas its straightening peak occurs at L19. This produces a much larger entropy--straightening gap than in the other categories ($-16$ vs. about $-4$ at the peak level), and the layer-wise correlation between prompt selectivity and straightening is also lower. This profile suggests that, in the evaluated code-generation benchmarks, prompt-selective access happens very early and is less tightly coupled to the later straightening signal. This is consistent with the main quality table, where the Code category remains relatively robust even under the aggressive $K=16$ cutoff. We therefore interpret Coding not as a failure case for the diagnostic, but as evidence that the required prefill-visible depth is task-dependent.

These diagnostics support an access-to-stabilization interpretation of the cutoff frontier, rather than a single-peak cutoff rule. For most non-coding categories, $K=16$ includes many attention-based peak layers but leaves little direct prompt-visible computation after selective prompt access and before the later stabilization region. $K=20$ covers more of this transition, but leaves little buffer beyond the observed straightening peak. $K=24$ retains the observed selection-to-stabilization interval for the broader benchmark suite with a small buffer, while $K=28$ retains the same interval more conservatively. Thus, we interpret SPEED-24+BoS and SPEED-28+BoS as broad operating points on the quality--efficiency frontier, not as universal per-task optima. Conversely, Coding suggests that some task families may tolerate shallower prefill-token visibility.

We emphasize that this analysis is diagnostic rather than causal. Straightening should not be interpreted as token independence; a safer interpretation is that hidden-state trajectories become more geometrically stabilized after selective contextual integration. Moreover, sample-level alignment between entropy minima and straightening peaks varies substantially across categories, so the analysis should be understood as category-level evidence rather than a per-example cutoff predictor. The diagnostics also do not imply that upper-layer computation is unnecessary: Decode-phase tokens remain full-depth in SPEED, and the SelfOnly ablation in Appendix~\ref{app:selfonly} separately shows that upper-layer Decode-token attention is important.

\section{Upper-layer Decode-token Attention Ablation}
\label{app:selfonly}

SPEED removes upper-layer KV states for prefill tokens, but it does not remove upper-layer attention among Decode-phase tokens. To test whether this Decode-token attention is necessary, we evaluate a SelfOnly diagnostic variant. Like SPEED, SelfOnly follows the same shallow-Prefill visibility rule. Unlike SPEED, upper-layer Decode-phase tokens do not attend to other Decode-phase tokens; they attend only to their own current position, optionally with a BoS anchor.

\begin{fixedtable}
\caption{
General capability for upper-layer Decode-token attention ablations.
SelfOnly removes upper-layer attention to other Decode-phase tokens and keeps only self-attention in upper layers, optionally with a BoS anchor.
}
\label{tab:selfonly-general}
\centering
\small
\setlength{\tabcolsep}{3.0pt}
\begin{tabular}{lcccccc}
\toprule
Method & Avg. & Know. & Reason. & Code & Math & Inst. \\
\midrule
Full-IT & \textbf{51.4} & 44.9 & 57.8 & 74.6 & \textbf{46.6} & \textbf{36.1} \\
IT-SPEED-28+BoS & 51.3 & 45.7 & 57.0 & 75.2 & \textbf{46.6} & 34.9 \\
SelfOnly-28+BoS & 50.0 & 43.0 & \textbf{58.0} & 73.5 & 44.3 & 34.0 \\
SelfOnly-28 & 50.0 & 42.9 & 57.9 & \textbf{75.4} & 43.5 & 34.1 \\
\midrule
IT-SPEED-24+BoS & 51.2 & \textbf{46.0} & \textbf{58.0} & \textbf{75.4} & 45.3 & 33.9 \\
SelfOnly-24+BoS & 47.2 & 40.6 & 55.3 & 71.0 & 41.5 & 31.3 \\
SelfOnly-24 & 47.5 & 41.5 & 48.5 & 71.7 & 42.4 & 31.6 \\
\bottomrule
\end{tabular}
\end{fixedtable}

The SelfOnly diagnostic shows that upper-layer Decode-token attention is not redundant. At $K=28$, SelfOnly-28+BoS reaches 50.0 average score, compared with 51.3 for IT-SPEED-28+BoS. The degradation is moderate and spread across categories: Knowledge drops from 45.7 to 43.0, Code from 75.2 to 73.5, Math from 46.6 to 44.3, and Instruction from 34.9 to 34.0, while Reasoning slightly increases from 57.0 to 58.0. Thus, preserving upper-layer feed-forward computation and a BoS anchor is not sufficient to fully match SPEED+BoS, but the degradation should not be interpreted as primarily math-specific.

At $K=24$, removing upper-layer Decode-token attention has a larger aggregate effect. SelfOnly-24+BoS reaches 47.2 average score, compared with 51.2 for IT-SPEED-24+BoS, with drops across Knowledge, Reasoning, Code, Math, and Instruction. This suggests that upper-layer Decode-token attention becomes more important when fewer lower layers retain direct prefill-token visibility. The anchor-free SelfOnly-24 variant is a broader stress test because it also removes the BoS anchor; its average score is similar to SelfOnly-24+BoS, but its category profile differs substantially, especially in Reasoning. We therefore use SelfOnly as a diagnostic of Decode-token attention rather than as evidence about the optimal anchor design.

\section{Repetition-loop Analysis}
\label{app:repetition-analysis}

Anchor-free SPEED can degrade generation not only by producing incorrect answers, but also by inducing suffix repetition loops. We therefore analyze suffix repetition loops as an additional diagnostic of generation stability. This analysis is separate from task accuracy and is intended to identify a specific failure mode caused by removing all full-depth prefill-side anchors.

For a prediction file with $n$ examples, we define
\begin{equation}
\mathrm{LoopRate}(\%) = 100 \times \frac{\mathrm{loop\_count}}{n},
\end{equation}
where $\mathrm{loop\_count}$ is the number of outputs flagged as suffix repetition loops.
We extract generated text from each prediction file, normalize whitespace and simple token boundaries, and tokenize the output with a lightweight regex tokenizer rather than the model tokenizer. We inspect only the final 256 tokens of each output. Within this tail window, we search for repeated token units of length 1 to 20 tokens, allowing up to 8 trailing tokens after the repeated suffix and allowing the final repeated unit to be partial. An output is flagged as a loop when the repeated suffix spans at least 12 tokens and repeats at least three times:
\begin{equation}
\mathrm{has\_loop}
=
\mathbf{1}\left[
\mathrm{loop\_tokens} \ge 12
\;\land\;
\mathrm{loop\_repeats} \ge 3
\right].
\end{equation}
This heuristic targets short exact suffix loops near the end of generation; it does not attempt to detect all semantic repetition or non-suffix repetition.

\begin{fixedtable}
\caption{
Suffix repetition loop rate (\%) across general-capability benchmarks.
Lower is better.
}
\label{tab:loop-rate}
\centering
\small
\setlength{\tabcolsep}{2.0pt}
\begin{tabular}{@{}lcccccccccccc@{}}
\toprule
Method & Avg. & MMLU & TQA & PopQA & BBH & CHE & CHE+ & GSM & DROP & MATH & IFEval & AE2 \\
\midrule
Full-IT ($K=32$)
& 0.4 & 0.1 & 0.0 & 0.0 & 0.5 & 0.0 & 0.0 & 0.1 & 0.0 & 2.0 & 1.3 & 0.7 \\
\midrule
IT-SPEED-28
& 1.4 & 0.1 & 0.0 & 7.6 & 1.1 & 0.0 & 0.0 & 1.6 & 0.0 & 3.2 & 1.7 & 0.5 \\
IT-SPEED-28+BoS
& 0.6 & 0.1 & 0.0 & 0.0 & 1.4 & 0.0 & 0.0 & 0.4 & 0.0 & 2.1 & 1.5 & 1.0 \\
\midrule
IT-SPEED-24
& 2.1 & 0.1 & 0.0 & 10.3 & 2.6 & 0.0 & 0.0 & 3.1 & 0.0 & 4.0 & 1.8 & 0.9 \\
IT-SPEED-24+BoS
& 0.7 & 0.1 & 0.0 & 0.0 & 1.2 & 0.0 & 0.0 & 0.5 & 0.0 & 2.3 & 2.0 & 1.4 \\
\midrule
IT-SPEED-20
& 0.8 & 0.7 & 0.0 & 0.1 & 1.4 & 0.0 & 0.0 & 0.7 & 0.0 & 3.8 & 1.5 & 0.7 \\
IT-SPEED-20+BoS
& 0.7 & 0.3 & 0.0 & 0.0 & 0.8 & 0.6 & 0.0 & 0.3 & 0.0 & 2.5 & 2.4 & 0.9 \\
\bottomrule
\end{tabular}
\end{fixedtable}

The results show that anchor-free SPEED increases suffix repetition loops relative to Full-IT, especially on PopQA and GSM. Adding a BoS anchor substantially reduces this failure mode without restoring upper-layer KV states for the full prefill sequence. Rows with missing values correspond to incomplete prediction files and are reported only for the available benchmark-level diagnostics.

\section{Long-context Length Robustness}
\label{app:length-robustness}

We further evaluate whether SPEED+BoS preserves performance across long prompt lengths. This analysis complements the aggregate downstream transfer results by grouping examples according to prompt length. TriviaQA represents naturally varying document lengths, while S-NIAH is a synthetic retrieval stress test with contexts extending to approximately 130K tokens.

\begin{figure}[!htbp]
    \centering
    \includegraphics[width=0.82\linewidth]{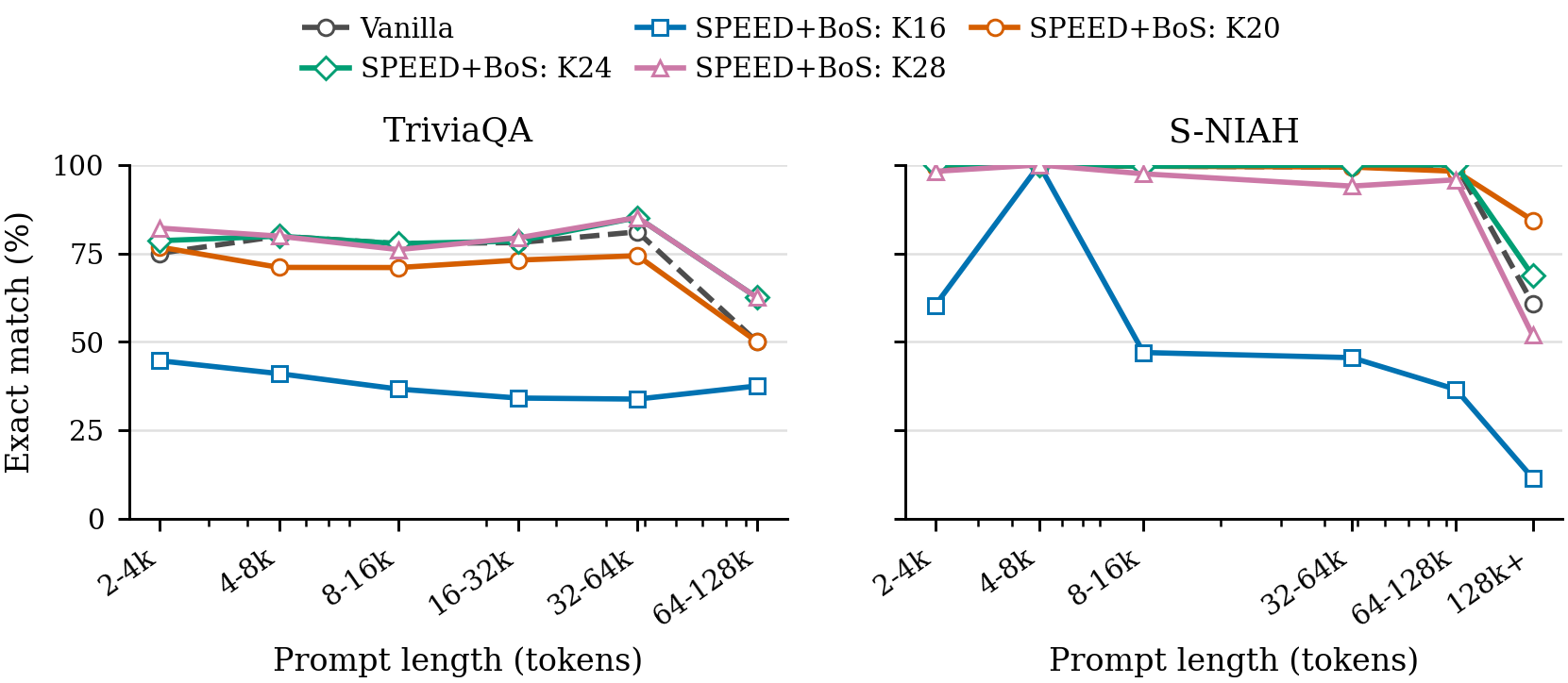}
    \caption{
    Exact match by prompt length on TriviaQA and S-NIAH.
    SPEED+BoS with moderate or conservative cutoffs remains competitive with Full-IT across long-context buckets, while aggressive $K=16$ degrades substantially.
    }
    \label{fig:length-robustness}
\end{figure}

We use this analysis to test whether SPEED+BoS can still exploit long prompts, not to rank cutoffs by intrinsic retrieval ability. Bucket-level S-NIAH scores may be affected by instance composition and evaluation variance.

\section{Task-adaptive and Compatibility Results}
\label{app:task-adaptive-compat}

\subsection{Downstream transfer without task-adaptive fine-tuning}
\label{app:downstream-transfer-full}

\begin{fixedtable}
\caption{
Downstream transfer results for instruction-tuned models without task-adaptive fine-tuning.
QA columns report EM/F1. CNN/DailyMail reports BERTScore F1~\citep{zhang2019bertscore}.
}
\label{tab:downstream-transfer-full}
\centering
\begin{tabular}{lccccc}
\toprule
Method & HotpotQA & TriviaQA & NQ & S-NIAH & CNN/DM \\
\midrule
Full-IT & 55.4 / 69.7 & 78.3 / 84.6 & 47.6 / 60.8 & 93.3 & 24.7 \\
PostHoc-SPEED-28 & 55.3 / 69.7 & 77.5 / 83.9 & 48.1 / 60.8 & 93.2 & 20.8 \\
PostHoc-SPEED-24 & 37.5 / 50.6 & 66.9 / 73.9 & 35.3 / 47.3 & 80.6 & 14.5 \\
IT-SPEED-28+BoS & 56.7 / 70.0 & 78.7 / 84.2 & 47.3 / 60.0 & 89.3 & 25.8 \\
IT-SPEED-24+BoS & 55.4 / 69.7 & 78.9 / 84.4 & 46.3 / 58.9 & 94.7 & 24.6 \\
IT-SPEED-20+BoS & 51.0 / 65.3 & 71.9 / 78.7 & 42.9 / 55.1 & 96.8 & 23.1 \\
IT-SPEED-16+BoS & 29.1 / 49.3 & 37.6 / 58.2 & 22.1 / 34.8 & 44.2 & 22.4 \\
\bottomrule
\end{tabular}
\end{fixedtable}

PostHoc-SPEED applies the SPEED visibility policy only at inference time to a model trained with full-depth prefill-token KV visibility. The large degradation at $K=24$ shows that stronger prefill truncation benefits from SPEED-aware adaptation. In contrast, the SPEED-aware BoS-anchored models remain competitive at moderate cutoffs, especially $K=24$ and $K=28$, supporting the controlled instruction-tuning results in the main paper.

\subsection{Task-adaptive fine-tuning}
\label{app:taskft-full}

\begin{fixedtable}
\caption{
Task-adaptive document QA and summarization results.
QA columns report EM/F1. CNN/DailyMail reports BERTScore F1~\citep{zhang2019bertscore}.
}
\label{tab:taskft-docqa-summary-full}
\centering
\begin{tabular}{lccccc}
\toprule
Method & HotpotQA & TriviaQA & NQ & S-NIAH & CNN/DM \\
\midrule
TaskFT-Full & 61.0 / 75.3 & 79.6 / 85.4 & 49.3 / 62.2 & 93.3 & 27.4 \\
TaskFT-SPEED-28+BoS & 60.9 / 75.3 & 80.9 / 86.1 & 50.2 / 62.6 & 92.9 & 27.8 \\
TaskFT-SPEED-24+BoS & 61.2 / 75.5 & 80.3 / 85.6 & 49.2 / 61.4 & 92.6 & 27.5 \\
TaskFT-SPEED-20+BoS & 60.2 / 74.1 & 78.5 / 84.2 & 48.2 / 60.7 & 98.9 & 28.1 \\
\bottomrule
\end{tabular}
\end{fixedtable}

\begin{fixedtable}
\caption{
Task-adaptive math and code transfer.
Math models are fine-tuned on Nemotron-Math; code models are fine-tuned on OpenCodeInstruct.
}
\label{tab:taskft-math-code-full}
\centering
\begin{tabular}{lcc}
\toprule
Method & MathBench & BigCodeBench \\
\midrule
TaskFT-Full & 50.4 & 23.7 \\
TaskFT-SPEED-28+BoS & 48.8 & 23.8 \\
TaskFT-SPEED-24+BoS & 53.0 & 24.2 \\
TaskFT-SPEED-20+BoS & 52.8 & 20.3 \\
\bottomrule
\end{tabular}
\end{fixedtable}

The task-adaptive results show that moderate SPEED+BoS cutoffs remain compatible with downstream adaptation. On document QA and summarization, $K=24$ and $K=28$ remain close to TaskFT-Full. On math and code, TaskFT-SPEED-24+BoS is competitive with or slightly above the full-depth task-adapted baseline in this setting. We treat these results as compatibility evidence rather than as the primary basis for the quality--efficiency frontier, since they use task-specific adaptation data and fewer model variants than the controlled instruction-tuning sweep.

\subsection{Off-the-shelf instruction-model LoRA pilot}
\label{app:offshelf-pilot}

This pilot tests whether SPEED-style adaptation can be applied starting from an off-the-shelf instruction-following checkpoint rather than only from the controlled Base-to-SFT pipeline. All adapted rows start from Llama-3.1-8B-Instruct and use one epoch of LoRA task adaptation on HotpotQA pseudo-labeled training examples.

\begin{fixedtable}
\caption{
Off-the-shelf instruction-model compatibility with lightweight SPEED adaptation.
Full-depth LoRA denotes full-depth adaptation under the same setup.
QA columns report EM/F1; S-NIAH reports exact match.
}
\label{tab:offshelf-pilot}
\centering
\small
\setlength{\tabcolsep}{3.5pt}
\begin{tabular}{lcccc}
\toprule
Method & HotpotQA & TriviaQA & NQ & S-NIAH \\
\midrule
Llama3.1 8B Instruct & 56.9 / 72.7 & 78.8 / 84.8 & 45.8 / 61.1 & 99.6 \\
Full-depth LoRA & 60.8 / 75.3 & 80.5 / 86.0 & 48.5 / 62.4 & 97.7 \\
OffShelf-FT-SPEED+BoS-28 & 58.7 / 73.4 & 81.3 / 86.5 & 47.9 / 61.5 & 97.0 \\
OffShelf-FT-SPEED+BoS-24 & 59.5 / 73.7 & 81.4 / 86.5 & 46.4 / 59.8 & 99.6 \\
OffShelf-FT-SPEED+BoS-20 & 59.4 / 73.5 & 81.1 / 86.4 & 45.4 / 58.7 & 96.1 \\
OffShelf-FT-SPEED+BoS-16 & 55.0 / 69.4 & 76.7 / 81.7 & 39.2 / 52.8 & 88.8 \\
\bottomrule
\end{tabular}
\end{fixedtable}

The results suggest that moderate cutoffs remain usable after lightweight task adaptation from an off-the-shelf instruction model. Since the adaptation data come from HotpotQA, the TriviaQA and S-NIAH results are useful transfer checks rather than direct measurements of fitting the adaptation task alone.

\section{Training Efficiency}
\label{app:training-efficiency}

Although our primary focus is inference, the same visibility policy can reduce the cost of prompt-heavy supervised adaptation when only lightweight trainable modules are updated. We therefore measure downstream LoRA fine-tuning efficiency under matched data order, effective batch size, optimizer, precision, hardware, activation-checkpointing policy, and gradient-accumulation setting across methods. Full fine-tuning did not show the same magnitude of wall-clock gain, so we treat LoRA adaptation efficiency as an auxiliary result rather than a primary claim.

\begin{fixedtable}
\caption{
Training efficiency in downstream LoRA fine-tuning.
All rows use one GPU. Effective tokens/sec excludes padding. Speedup is computed from GPU-hours relative to Vanilla.
}
\label{tab:lora-training-efficiency}
\centering
\small
\begin{tabular}{lccccc}
\toprule
Method & $K$ & GPU-hours $\downarrow$ & Eff. tok/s/GPU $\uparrow$ & Peak GiB $\downarrow$ & Speedup \\
\midrule
Vanilla & 32 & 8h19m & 2213.8 & 63.4 & $1.00\times$ \\
SPEED-28+BoS & 28 & 7h22m & 2499.7 & 62.2 & $1.13\times$ \\
SPEED-24+BoS & 24 & 6h26m & 2863.1 & 61.6 & $1.29\times$ \\
SPEED-20+BoS & 20 & 5h25m & 3395.1 & 61.2 & $1.54\times$ \\
SPEED-16+BoS & 16 & 4h13m & 4366.9 & 60.8 & $1.97\times$ \\
\bottomrule
\end{tabular}
\end{fixedtable}

The main benefit is wall-clock throughput rather than peak-memory reduction: peak memory changes only modestly from 63.4 GiB to 61.6 GiB at $K=24$, whereas effective token throughput increases from 2213.8 to 2863.1 tokens/s/GPU. This pattern is consistent with SPEED reducing prefill-token layer computation while leaving optimizer state, LoRA parameters, and much of the training memory footprint unchanged.

\section{Broader Impacts}
\label{app:broader-impacts}

SPEED aims to reduce the compute and memory cost of long-context language-model inference. A positive impact is that such efficiency improvements can reduce serving cost, lower energy use per request, and make long-context LLM systems more accessible to researchers and practitioners with limited compute. A potential negative impact is that cheaper long-context generation may also lower the cost of misuse, including large-scale automated text generation, processing of sensitive documents, or deployment in settings where model errors can affect users. SPEED does not introduce application-specific safety mechanisms, so deployments should follow the safety, privacy, and usage restrictions of the underlying model and application domain.

\end{document}